\documentclass{article}

\usepackage[final,main]{neurips_2026}

\usepackage[utf8]{inputenc}
\usepackage[T1]{fontenc}
\usepackage{hyperref}
\usepackage{url}
\usepackage{booktabs}
\usepackage{amsfonts} 
\usepackage{amsmath}       
\usepackage{amssymb,mathtools}
\usepackage{nicefrac}
\usepackage{microtype}
\usepackage{amsthm}
\usepackage{xcolor}
\usepackage{graphicx}
\usepackage{subcaption}
\usepackage{bbm}
\usepackage{algorithm}
\usepackage{algpseudocode}
\usepackage{enumitem}
\usepackage{tabularx}
\usepackage{array}
\usepackage{multirow}
\usepackage{float}
\usepackage{wrapfig}
\usepackage{bbm}
\usepackage{colortbl}

\newtheorem{claim}{Claim}

\usepackage[most,skins,theorems]{tcolorbox}   
\usepackage{longtable}                          
\usepackage{tabularx}                           
\usepackage{wrapfig}                            
\usepackage{multirow}                           
\usepackage{makecell}                           
\usepackage{enumitem}                           
\usepackage{algorithm}                          
\usepackage{algpseudocode}                      
\usepackage{booktabs}                           

\hypersetup{
  colorlinks=true,
  linkcolor=black,
  citecolor=black,
  urlcolor=blue,
  pdftitle={A Subgoal-driven RL Framework for Improving Long-Horizon Web Agents}
}

\title{A Subgoal-driven RL Framework for Improving Long-Horizon Web Agents}
 
\author{
    \textbf{Taiyi Wang}\textsuperscript{1},
    \textbf{Sian Gooding}\textsuperscript{1},
    \textbf{Florian Hartmann}\textsuperscript{1},
    \textbf{Oriana Riva}\textsuperscript{1} and
    \textbf{Edward Grefenstette}\textsuperscript{1}
    \\
    \textsuperscript{1}Google DeepMind
}
\begin{document}
\maketitle
 
\begin{abstract}
Large language model (LLM)-based agents have emerged as powerful autonomous controllers for digital environments, spanning mobile interfaces, operating systems, and web browsers. Web navigation, for example, demands handling dynamic content and long action sequences, making it a particularly complex task. Existing LLM-backed agents exhibit weakened long-horizon planning abilities during RL fine-tuning, where sparse and delayed rewards from terminal Outcome Reward Models (ORMs) make it difficult for agents to identify the actions that lead to success, preventing them from sustaining coherent reasoning over extended tasks. We address this with two contributions: (1) a validated subgoal generation procedure that produces reliable, monotonically calibrated progress signals between the initial state and the terminal ORM reward; and (2) MiRA (\underline{Mi}lestoning your \underline{R}einforcement Learning Enhanced \underline{A}gent), an RL training framework using dense, milestone-based reward signals via a learned potential critic. Starting from the open Gemma3-12B base ($6.4\%$), MiRA enables the agent to reach $43.0\%$ on the WebArena-Lite benchmark. This performance surpasses proprietary systems such as GPT-4-Turbo ($17.6\%$) and GPT-4o ($13.9\%$), as well as the previous open-model state of the art, WebRL ($38.8\%$). Our findings demonstrate that milestone-based reward shaping significantly boosts an agent's long-horizon abilities, paving the way for more robust, general-purpose autonomous systems.
\end{abstract}
\section{Introduction}
\label{sec:intro}
 
\begin{figure}[htbp]
    \centering
    \includegraphics[width=\linewidth]{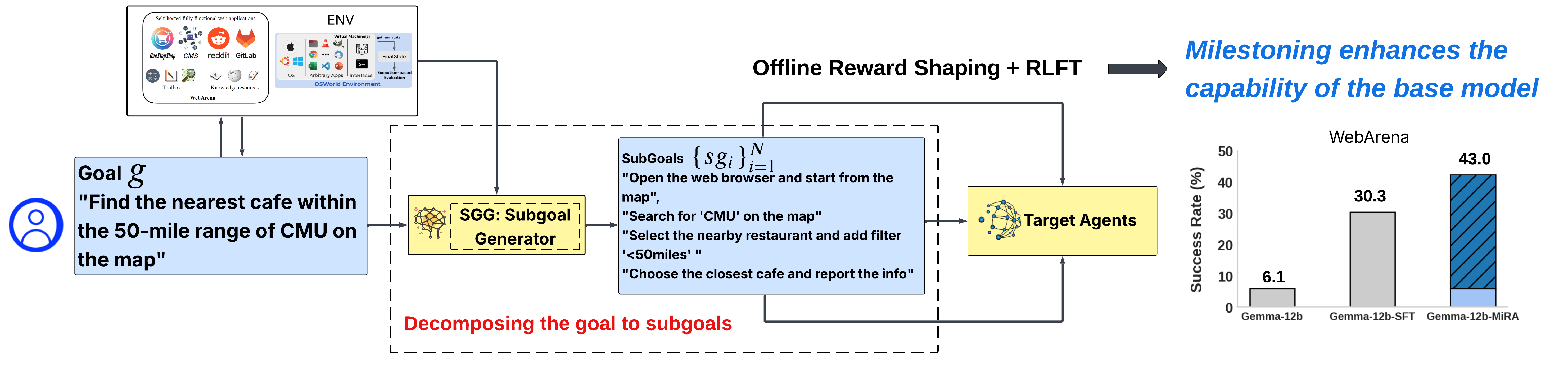}
    \caption{\textbf{Overview of MiRA.} A high-level task goal is decomposed into a sequence of subgoals by a Subgoal Generator (SGG). These subgoals drive reward shaping during RL fine-tuning of the target agent.}
    \label{fig:intro}
\end{figure}
 
LLM-based agents have gained significant traction as autonomous interfaces for real-world digital environments, spanning mobile device automation~\citep{yang2023appagent,christianos2024lightweight,rawles2024androidworld}, OS control~\citep{xie2024osworld,zhang2025ufo}, and web navigation~\citep{wang2024distrl,qi2024webrl,deng2023mind2web}. Among these, \textbf{web navigation} serves as a particularly rigorous testbed: even frontier proprietary systems that perform strongly on aggregate UI-control tasks suffer marked performance drops on open-ended, long-horizon web benchmarks~\citep{zhang2024large,GoogleDeepMind_Gemini2.5_2025}. Smaller open-source models face an even wider gap, typically relying on supervised fine-tuning (SFT) to bridge it. However, SFT is limited by static data and poor generalization, motivating a shift toward \textbf{reinforcement learning~(RL)} for long-horizon optimization~\citep{bai2024digirl,wang2024distrl}. Realistic benchmarks such as WebArena~\citep{zhou2023webarena} and Mind2Web~\citep{deng2023mind2web} further underscore the challenge: current agents---proprietary and open-source alike---exhibit sharp performance degradation as task complexity increases.
 
Our empirical analysis further quantifies this degradation. On the WebArena-Lite benchmark~\citep{liu2024visualagentbench,koh2024visualwebarena}, proprietary models such as Gemini-2.5-Pro exhibit ``mid-task stuck'' behaviors in nearly 50\% of evaluation trajectories, while SFT-trained open models like Gemma-12B-SFT still fail to progress in over 30\% of cases. The same deficiency dominates error distributions in prior RL-trained agents such as WebRL~\citep{qi2024webrl}, where high-level goal instability leads to divergence from the target objective. Collectively, these observations highlight that current models---regardless of scale or fine-tuning---lack the robust internal planning and progress-awareness required to sustain reasoning over extended interactions, suggesting that a key bottleneck is \emph{long-horizon credit assignment under sparse rewards}.

 
A promising direction is to introduce \textbf{explicit subgoals} as intermediate checkpoints that guide the agent toward the final objective. However, integrating such milestoning into agents raises two key challenges:
\textbf{(C.1)} \emph{Where do subgoals come from, and how reliable are they?}
\textbf{(C.2)} \emph{How can intermediate rewards be embedded in RL training to improve credit assignment while controlling the auxiliary signal's influence on the true task objective?}
 
To address these challenges, we propose \textbf{MiRA} (\underline{Mi}lestoning your \underline{R}einforcement Learning Enhanced \underline{A}gent), a subgoal-oriented RL fine-tuning framework. As illustrated in Figure~\ref{fig:intro}, our system decomposes high-level goals into structured subgoals and uses them to provide dense, potential-based reward shaping during training. Fundamentally, our approach follows a simple principle: \emph{``If the final goal is difficult to reach directly, increasing the probability of reaching meaningful intermediate milestones helps.''} Our contributions are as follows:
 
\begin{enumerate}[leftmargin=*,itemsep=2pt]

    \item \textbf{Solid Subgoal Generation.} We present an iterative in-context mutation strategy that generates reliable, semantically grounded subgoals, calibrated against final-task success to serve as dense reward signals as well as progress indicators for RL fine-tuning (\S\ref{sec:sg_gen}).
 
    \item \textbf{MiRA: Milestone-Based RL Fine-Tuning.} We introduce a dual-critic RL architecture combining a goal-conditioned value critic with a learned potential critic for dense, subgoal-driven reward shaping, trained via an iterative failure-driven curriculum. Applied to Gemma3-12B, MiRA substantially outperforms prior open-source and proprietary baselines on the verified subset of WebArena (\S\ref{sec:mira_rl}).
\end{enumerate}

\section{Related Work}
\label{sec:related}
 
\subsection{GUI Agents and Reinforcement Learning Fine-Tuning}
\label{subsec:gui_rl}
 
LLM-based agents for GUI control can be grouped into three paradigms: \emph{prompting-based}, \emph{imitation-based}, and \emph{RL-based}. Prompting-based agents steer frozen models through structured instructions and tool use, achieving strong zero-shot performance but limited adaptability. Imitation-based agents rely on SFT over demonstrations, which improves task alignment but depends on static data and fails to teach recovery from errors~\citep{zhai2024fine}. This brittleness has motivated a shift toward reinforcement fine-tuning, where agents learn from active interaction. However, even with RL, adapting to complex workflows remains difficult due to the sparsity of feedback signals in multi-step environments. This challenge persists across the spectrum of GUI-based agents, including mobile controllers~\citep{bai2024digirl,yang2023appagent,wang2024distrl}, OS assistants~\citep{zhang2025ufo,xie2024osworld}, and web agents. In the web domain, the WebRL framework~\citep{qi2024webrl} applies a self-evolving curriculum to address training-task scarcity, feedback sparsity, and distribution drift. Despite these advances, rewards in web navigation are often binary after many interactions, and RL-based agents still show steep performance drops as task length increases.
 
In hierarchical settings, an LLM planner produces intermediate sub-goals that a goal-conditioned policy executes~\citep{wang2023describe,zhang2025goal}, allowing semantic goal decomposition and pruning of irrelevant actions. However, even such planning architectures face critical limitations: tasks exceeding ${\sim}10$ steps often fail due to impaired sub-goal coherence and drift from the original aim~\citep{qian2025webgrapheval,shlomov2025benchmarks}. Recent efforts have introduced Process Reward Models (PRMs) to mitigate these failures~\citep{xi2025agentprm,cui2025process}. Unlike Outcome Reward Models that provide sparse feedback only upon task completion, PRMs offer dense, step-by-step supervision. For instance, Web-Shepherd~\citep{chae2025web} leverages checklist-style sub-goal verification, while AgentPRM~\citep{xi2025agentprm} introduces a dual-scoring mechanism measuring both ``promise'' and ``progress.'' However, learned PRMs often suffer from high inference overhead and susceptibility to reward over-optimization. Our approach instead utilizes explicit milestones as rigid semantic checkpoints, retaining continuous progress tracking while ensuring verifiable reliability.

Prompting-based web agents such as WebVoyager~\citep{he2024webvoyager} and
AgentOccam~\citep{yang2025agentoccam} improve navigation through multimodal
interaction or observation--action alignment, while SiRA~\citep{deng2025sira}
uses world-model simulation for inference-time planning. In contrast, MiRA
focuses on \emph{training-time} credit assignment: semantic milestones are
converted into progress targets for RL rather than a fixed inference-time plan.

\vspace{-5pt}
\subsection{Goal-Conditioned Reinforcement Learning}
\label{subsec:gcrl}
 
Sequential web tasks naturally align with goal-conditioned reinforcement learning (GCRL)~\citep{kaelbling1993learning}, where the agent optimizes cumulative reward conditioned on a task goal. A major difficulty is \emph{reward sparsity}, which hinders efficient credit assignment. Hindsight Experience Replay (HER)~\citep{andrychowicz2017hindsight} addresses this by reinterpreting failed trajectories as successes for alternative goals, but assumes Markovian rewards that limit applicability to long-horizon web navigation. Recent work has pivoted toward on-policy GCRL: GCPO~\citep{gong2024goal} leverages self-curriculum learning for non-Markovian rewards, while WebAgent-R1~\citep{wei2025webagent} demonstrates that end-to-end on-policy RL with ``thinking'' steps can surpass imitation baselines.
 
Explicit subgoal scaffolding remains critical for exploration. VSC-RL~\citep{wu2025vsc} utilizes subgoal-conditioned learning to boost sample efficiency but struggles to balance intermediate completion with final-goal optimality. One Subgoal at a Time~\citep{guo2025onesubgoal} decomposes formally specified temporal-logic tasks into reach--avoid objectives, whereas MiRA learns reward supervision from natural-language web milestones. Our milestones serve as \emph{auxiliary} training signals; unlike classical PBRS, their use does not guarantee an unchanged optimal policy.
 
\vspace{-5pt}


\section{Preliminary and Motivation}
\label{sec:preliminary}
 
\subsection{Problem Formulation}
\label{sec:problem_formulation}
 
We formulate the web navigation task as a finite-horizon Partially Observable Markov Decision Process (POMDP), defined by the tuple $\mathcal{M} = \langle \mathcal{S}, \mathcal{A}, \Omega, \mathcal{T}, \mathcal{O}, R, H \rangle$. Here, $\mathcal{S}$ represents the latent environment state (e.g., server-side databases and hidden DOM elements), which is inaccessible to the agent. At each timestep $t$, the agent receives a partial observation $o_t \in \Omega$ (comprising the rendered HTML, screenshot, and task instruction $g$) governed by the observation function $\mathcal{O}(o_t \mid s_t)$.
 
The agent selects a discrete action $a_t \in \mathcal{A}$ (e.g., clicking, typing, scrolling) according to a policy $\pi(a_t \mid o_{\leq t})$ conditioned on the history of observations. Following the action, the environment transitions to the next state $s_{t+1}$ governed by the implicit dynamics $\mathcal{T}(s_{t+1} \mid s_t, a_t)$. The episode terminates either when the task is successfully verified or when the number of interactions reaches the horizon $H$, with the objective of maximizing the expected cumulative reward.
 
\noindent \textbf{State Representation.}
In web navigation tasks, the instantaneous observation alone is insufficient to characterize progress, because many actions depend on prior interactions (e.g., previously opened panels, typed queries, or navigation paths).
Since the latent environment state is inaccessible, we use the action--observation history as the agent's working state representation, and with slight abuse of notation we also denote it $s_t$:
$s_t = [a_1, a_2, \ldots, a_{t-1},\, o_t]$,
where $o_t$ denotes the current DOM tree, textual caption, or multimodal screenshot representation.

\noindent \textbf{Reward and Objective.}
The environment provides a sparse binary reward signal: $r_t(s_t, a_t, g) = \mathbbm{1}[s_{t+1} \in \mathcal{S}_g]$, where $\mathcal{S}_g$ denotes the set of states that satisfy the goal $g$. In practice, this reward is computed by an \emph{automatic LLM-as-Judge} that evaluates goal satisfaction using the task instruction, action history, and final-state information. We adopt the open-source ORM from WebRL~\citep{qi2024webrl} as the final goal-level checker, and Gemini-2.5-Pro for subgoal-level checking (prompts in Appendix~\ref{sec:apdx_exp}). The learning goal is to find a policy $\pi$ that maximizes the expected discounted return
$J(\pi) = \mathbb{E}_{\tau \sim \pi} \big[ \sum_{t=0}^{H} \gamma^t r_t \big]$, where $\gamma \in [0,1]$ is the discount factor.
 
\subsection{Potential-Based Reward Shaping (PBRS)}
\label{sec:pbrs}

In sparse-reward settings such as web navigation, credit assignment is difficult.
Classical Potential-Based Reward Shaping (PBRS)~\citep{ng2003shaping}
augments rewards by
$\tilde r_t = r_t + \gamma\Phi(s_{t+1}) - \Phi(s_t)$.
With the standard assumptions on state-based potentials and terminal handling,
this construction preserves optimal policies; it is also related to potential
initialization of the $Q$-function~\citep{wiewiora2003potential}
(proof for the classical rule in Appendix~\ref{sec:apdx_proof}).
MiRA is \emph{inspired} by potential differences, but deliberately uses
$\Phi(s_{t+1})-\Phi(s_t)$ without the next-state discount and trains its
outcome critic separately. Consequently, the classical policy-invariance
result does \emph{not} apply to the implemented MiRA objective.

\subsection{Motivation: Failure Analysis and Why Milestones Matter}
\label{sec:motivation}

Current web agents fail predominantly due to \emph{mid-task stagnation}: our automated failure analysis on the full WebArena benchmark~\citep{zhou2023webarena} (812 tasks across 5 domains, more details in Appendix~\ref{sec:apdx_failure}) reveals that \emph{Get Stuck Midway}---where agents enter non-productive action loops---accounts for 42--49\% of all failures across base models, their SFT variants, and even proprietary systems like Gemini-2.5-Pro. Agents that fail midway are often close to the solution but lack the means to break an incorrect loop. Notably, this mid-task stagnation mode dominates uniformly across open-source backbones, their SFT variants, and the specialized WebRL agent, indicating a structural deficiency rather than a capacity gap. This suggests that explicit subgoal signals---which inject intermediate supervision precisely where agents stall---would yield the largest gains, directly motivating our milestone-based approach to long-horizon credit assignment.

 


\section{Subgoal-Oriented Web Agent}
\label{sec:method}
 
Given the analysis above, we hypothesize that \emph{subgoal setting serves as a critical mechanism for strengthening long-horizon reasoning and stabilizing web agent performance}. We decompose our methodology into two components: generating reliable subgoals (addressing \textbf{C.1}) and enhancing RL fine-tuning through subgoal-driven reward shaping (addressing \textbf{C.2}).


\subsection{Subgoal Generation}
\label{sec:sg_gen}
 
\begin{wrapfigure}{r}{0.55\textwidth}
    \centering
    \vspace{-10pt}
    \includegraphics[width=\linewidth]{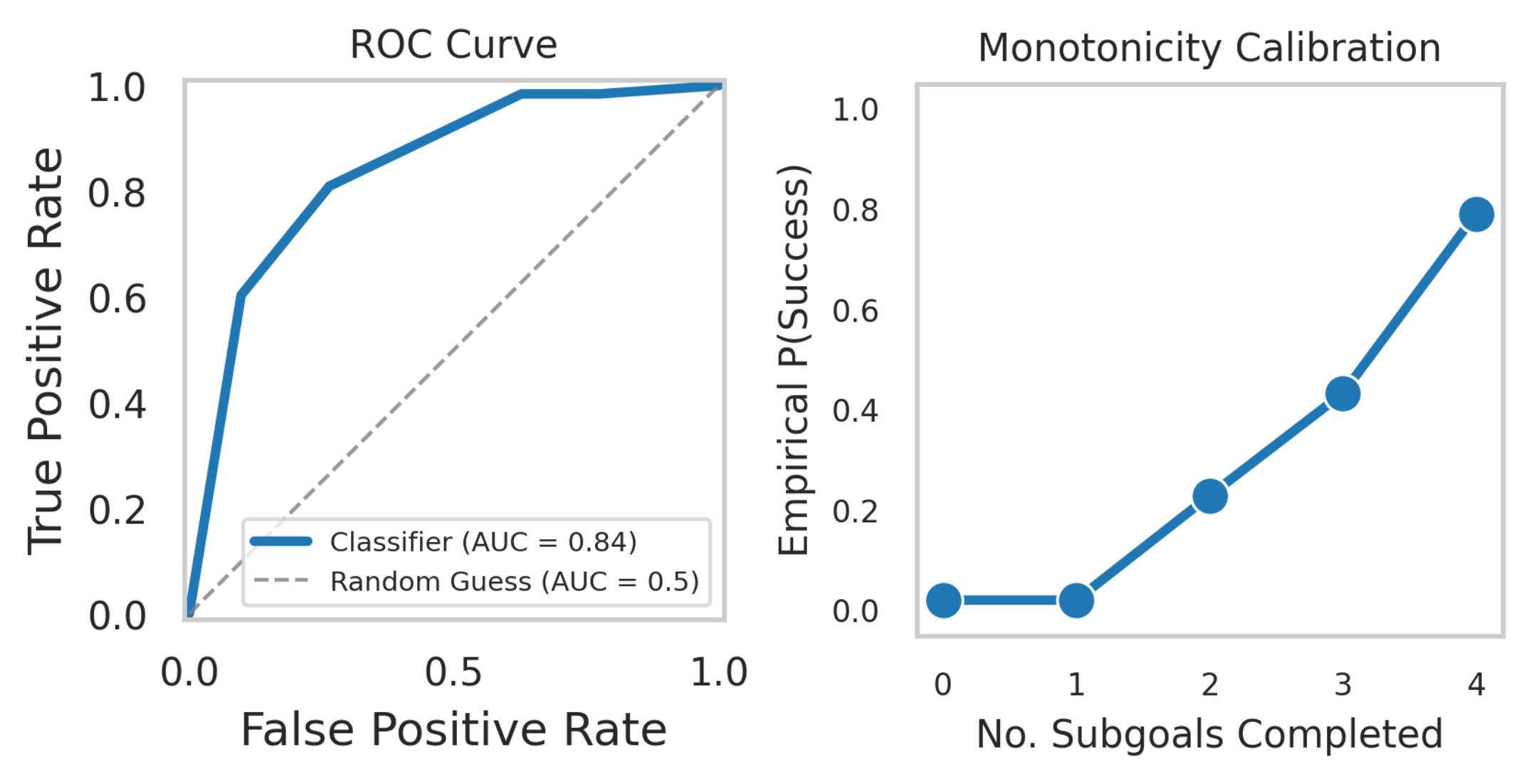}
    \caption{\textbf{Subgoal quality validation.} \textbf{(Left)} ROC curve using the fraction of completed subgoals as a score yields AUROC of 0.84. \textbf{(Right)} Monotonicity calibration shows strictly increasing $P(y{=}1 \mid m)$ as more subgoals $m$ are completed (Kendall's $\tau = 0.4585$, $p < 0.001$).}
    \vspace{-15pt}
    \label{fig:subgoal_validation}
\end{wrapfigure}
 
To effectively guide web agents, we generate subgoals that bridge the gap between abstract user intents and precise step-by-step actions. Achieving this requires ensuring both reliability and proper granularity, which necessitates a fundamental understanding of website-specific constraints and features. Representative examples of generated subgoals for each website category are provided in Appendix~\ref{sec:apdx_exp}.
 
\paragraph{Methodology.}
We leverage Gemini-2.5-Pro to generate subgoals given a high-level task description and the initial web state. To ensure robustness and generalization, we employ an iterative in-context learning strategy. Specifically, we curate a diverse set\footnote{2 examples for each website category} of few-shot demonstrations mapping intents to logical intermediate milestones. To mitigate positional bias and overfitting to specific action sequences, we randomize the distribution of these few-shot examples during prompting. This shuffling or mutation ensures the model attends to the \textit{semantics} of the task rather than memorizing rigid execution patterns. We iteratively optimize the generation context until convergence.
 
\paragraph{Evaluation of subgoal quality.}
To validate utility, we analyze the correlation between the generated subgoal completion vectors $\mathbf{z}_i \in \{0,1\}^K$ and the ground-truth final success $y_i \in \{0,1\}$ across a validation set of agent traces\footnote{Agent is trained with MiRA; base model is Gemma3-12B.}. Under the \textit{Exact Equivalence} regime, we observe an F1 of 0.6847 (Precision 0.7917, Recall 0.6032), indicating that completing all subgoals is a strong but not necessary condition for success. Under the \textit{Graded Agreement} regime, defining a progress score $s_i = \frac{1}{K}\sum_{k=1}^K z_{i,k}$, we find an AUROC of \textbf{0.84} (Figure~\ref{fig:subgoal_validation}, left) and a strictly monotonic relationship between the number of completed subgoals and the empirical success probability, confirmed by Kendall's $\tau = 0.4585$ ($p < 0.001$; Figure~\ref{fig:subgoal_validation}, right). These results support using graded milestone completion as an informative, though imperfect, auxiliary signal. Exact agreement is not required: a successful agent may take an alternative valid path or miss a checker-specific milestone, while completing the milestones alone may not ensure a correct final response. Executor and granularity analyses are given in Appendix~\ref{sec:apdx_ablation}.


\subsection{MiRA: Agent RL Fine-Tuning with Subgoals}
\label{sec:mira_rl}
 
\begin{figure}[t]
    \centering
    \includegraphics[width=\textwidth]{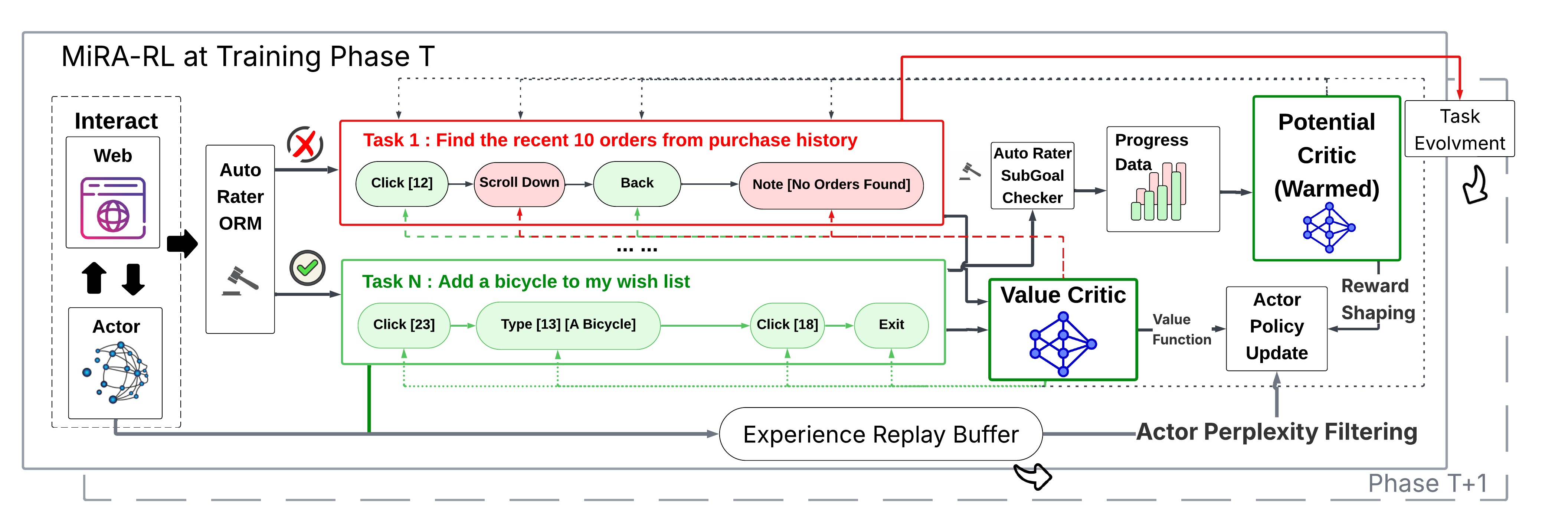}
    \caption{\textbf{The MiRA-RL training pipeline.} Trajectories are evaluated by an Auto Rater (binary success) and a SubGoal Checker (intermediate progress). Two critics are trained: a \textbf{Value Critic} $V_\phi$ for final success and a \textbf{Potential Critic} $P_\psi$ for progress. The actor is updated using shaped rewards, stabilized by perplexity filtering.}
    \label{fig:mira_workflow}
    \vspace{-10pt}
\end{figure}

To incorporate subgoal-level feedback and provide dense reinforcement signals, MiRA extends the value estimation framework of prior work~\citep{qi2024webrl,bai2024digirl,wang2024distrl}. As illustrated in Figure~\ref{fig:mira_workflow}, the training pipeline introduces a dual-critic architecture: a standard \emph{goal-conditioned value critic} $V_\phi(s,g)$ trained on binary outcomes (final success/failure), and a novel \emph{potential critic} $P_\psi(s,g)$ that acts as a dense \textbf{progress model} driven by a subgoal checker.
 
\begin{figure}[t]
    \centering
    \includegraphics[width=\textwidth]{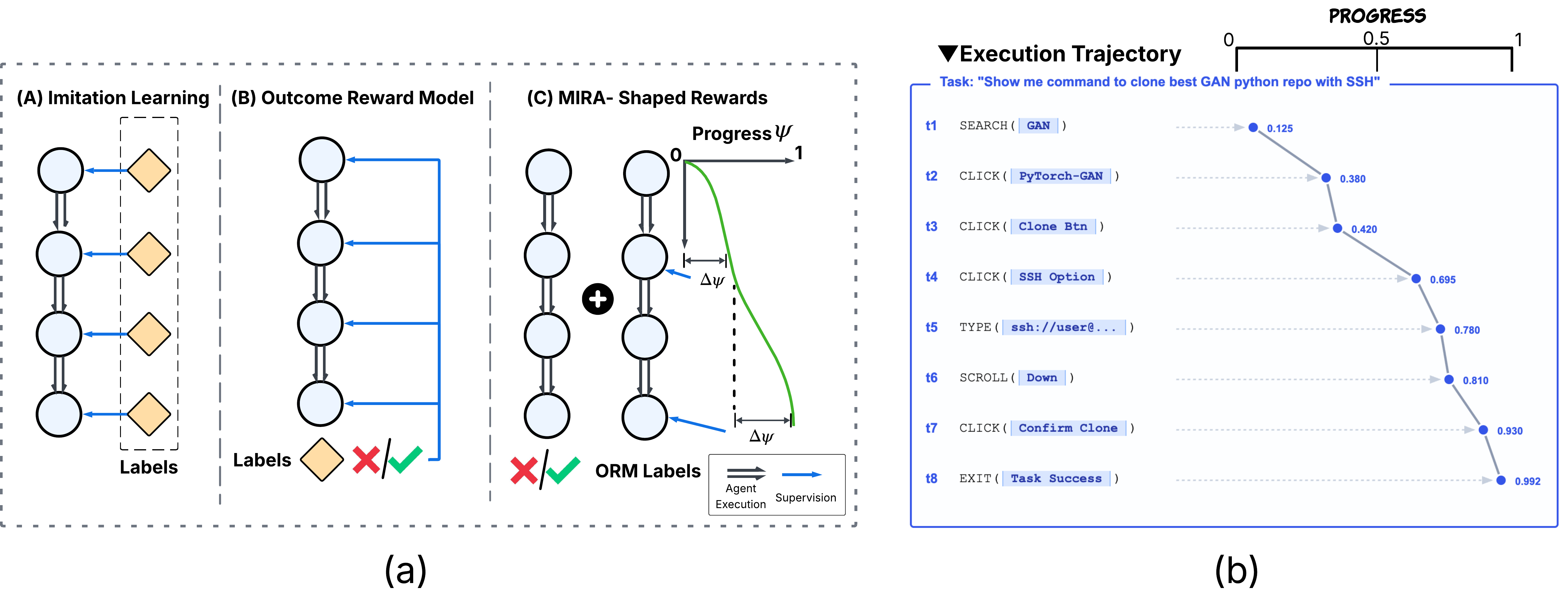}
    \caption{\textbf{Potential-based reward shaping.} \textbf{(Left)} Comparison of supervision signals: (A) Imitation learning matches exact steps; (B) ORM provides sparse terminal feedback; (C) MiRA uses shaped rewards via a learned potential function. \textbf{(Right)} Predicted progress score $\psi$ increases monotonically as the agent completes subgoals, generating dense $\Delta\psi$ rewards at every timestep.}
    \label{fig:potential_concept}
\end{figure}

\paragraph{Progress labeling via subgoal completion.}
For each trajectory and task goal $g$, we obtain a binary subgoal-completion vector at each step, $\mathbf{z}_t = (z_{t,1}, \dots, z_{t,K}) \in \{0,1\}^K$, where $z_{t,k}=1$ indicates the $k$-th subgoal has been achieved\footnote{Only positive traces are sampled.}. This allows the agent to maintain an explicit belief state of its progress, represented by the completion vector $\mathbf{z}_t$---a signal that can also be leveraged at inference time for dynamic re-planning. Key steps are defined where the cumulative count $c_t$ increases ($c_t > c_{t-1}$). To obtain smooth supervision, we assign normalized progress labels $p_t^\ast \in [0,1]$ through linear interpolation between key steps (Figure~\ref{fig:potential_concept}, right). Let $t_j$ and $t_{j+1}$ be the steps where the $j$-th and $(j{+}1)$-th subgoals complete. For any $t\in[t_j,t_{j+1}]$:
\begin{equation}
\label{eq:progress_interp}
    \alpha_t=\frac{t-t_j}{\,t_{j+1}-t_j\,},\quad
    p_t^\ast = (1-\alpha_t)\,\frac{j}{K} + \alpha_t\,\frac{j+1}{K}.
\end{equation}
 
\noindent \textbf{Example.} Consider a trajectory with $K=3$ subgoals completing at $t_1=2$, $t_2=4$. The final subgoal's logic check passes at $t=6$, but the agent performs concluding actions until termination at $T=9$. \emph{Segment~1} ($t \in [0,2]$): progress ramps $0 \to 0.33$. \emph{Segment~2} ($t \in [2,4]$): $0.33 \to 0.66$. \emph{Segment~3} ($t \in [4,9]$): we anchor the last subgoal to the trajectory end $T=9$ rather than $t=6$, so progress ramps $0.66 \to 1.0$ over $[4,9]$, ensuring the potential critic learns a monotonically increasing value through the final verification steps.
 
\paragraph{Potential critic for dense shaping.}
We train the potential critic $P_\psi(s,g)\in[0,1]$ to regress onto these progress labels. To construct a practical progress model, we augment a pretrained LLM with an MLP head and sigmoid activation. We use Llama3-8b (WebRL) and a vanilla-RL agent to perform exploratory rollouts across 1{,}237 tasks, post-process the trajectories using a subgoal checker to derive dense progress labels, and fine-tune Gemma-12B in a single supervised stage. Each state $s_t = [a_1, \ldots, a_{t-1}, o_t]$ captures the full action--observation history, enabling the critic to learn from the stable longest-common subsequences of actions that characterize successful trajectories. We then use this pre-trained network as a warm-up model and continue fine-tuning during RL. The critic is optimized via:
\begin{equation}
\label{eq:potential_regression}
\mathcal{L}_P(\psi) = \mathbb{E}_{(s_t,g,p_t^\ast)} \Big[ \| P_\psi(s_t,g) - p_t^\ast \|^2 \Big].
\end{equation}

\paragraph{Auxiliary reward shaping.}
During RL, the potential critic provides an auxiliary reward
$r'_t = r_t + \alpha\bigl[P_\psi(s_{t+1},g)-P_\psi(s_t,g)\bigr]$,
corresponding to the $\Delta\psi$ arrows in Figure~\ref{fig:potential_concept}.
The learned potential is supervised on verified successful traces, whereas
the outcome critic $V_\phi$ is trained exclusively on the terminal ORM label.
The resulting actor target combines outcome-based TD estimation with a shaped
Monte-Carlo return (Eq.~\ref{eq:advantage}). This is a \emph{success-anchored
surrogate}, not a policy-invariant PBRS advantage: the missing $\gamma$
in the potential difference and the terminal-only critic rule out applying
the classical guarantee~\citep{ng2003shaping,wiewiora2003potential}.
The terminal ORM still defines task success, but auxiliary shaping can influence
the learned policy. We therefore select a moderate shaping weight $\alpha$
empirically; larger weights degrade success rate (Appendix~\ref{sec:apdx_ablation}).


\paragraph{Goal-conditioned value critic.}
The primary value critic $V_\phi(s,I)$ models the probability that the trajectory passing through state $s$ ultimately succeeds on instruction $I$. We train it via binary cross-entropy against the terminal outcome label $r \in \{0,1\}$ assigned to each trajectory by the ORM (i.e., all transitions in a successful trajectory share label $r{=}1$, and all transitions in a failed trajectory share $r{=}0$):
\begin{equation}
\label{eq:critic_value}
\mathcal{L}_V(\phi) = -\mathbb{E}_{(s,I,r)\sim\mathcal{D}} \!\left[ r \log V_\phi(s,I) + (1{-}r)\log(1{-}V_\phi(s,I)) \right].
\end{equation}
Following \citet{farebrother2024stop,qi2024webrl}, we adopt this classification-based formulation to handle the sparse, binary nature of terminal rewards.

\paragraph{Actor optimization.}
We use a KL-regularized, log-ratio regression update motivated by Policy Mirror
Descent~\citep{abbasi2019politex,schulman2017equivalence}. For a conventional
regularized policy-improvement step, the target takes the exponential-tilting
form $\pi^*(a|s,I)\propto\pi_{\mathrm{ref}}(a|s,I)
\exp(A^*(s,a,I)/\beta)$ (Appendix~\ref{sec:apdx_RL_adv}).
In MiRA, however, the transition-level target is the surrogate defined below,
not an exact advantage for a single shaped MDP. The actor is fitted by MSE
regression on log-probability ratios:

\begin{equation}
\label{eq:mse_objective}
\mathcal{L}_\pi(\theta) = \mathbb{E}_{\nu}\!\left[\!\left(\beta\log\tfrac{\pi_\theta(a|s,I)}{\pi_{\text{ref}}(a|s,I)}-A_t^{\text{shaped}}\right)^{2}\!\right],
\end{equation}
where $A_t^{\text{shaped}}$ is a mixed TD--Monte-Carlo \emph{surrogate target} rather than an unbiased advantage estimator:
\begin{equation}
\label{eq:advantage}
A_t^{\text{shaped}} = \lambda \big(r'_t + \gamma V_\phi(s',g) - V_\phi(s,g)\big) + (1{-}\lambda)\big(G_t - V_\phi(s,g)\big), \quad G_t = \textstyle\sum_{u=t}^{T} \gamma^{u-t} r'_u.
\end{equation}
The objective permits regression on replay samples $\nu$ and adjusts log-probability ratios in both directions. The reference-policy term regularizes these updates, but it does not itself establish monotonic policy improvement or policy invariance. An exact decomposition of this success-anchored surrogate is given in Appendix~\ref{sec:apdx_proof}. The full procedure is in Algorithm~\ref{alg:mira-rl} (Appendix~\ref{sec:apdx_algo}).

\paragraph{Iterative curriculum refinement.}
A key limitation of static off-policy RL is vulnerability to local optima: once an agent saturates on a fixed task set, it ceases to generalize~\citep{qi2024webrl}. To address this, we wrap the MiRA inner loop (Algorithm~\ref{alg:mira-rl}) in an \emph{iterative curriculum} that alternates between environment interaction and offline policy updates. In each phase, the current policy collects rollouts, which are annotated with subgoal progress labels via a Gemini-2.5-Pro subgoal checker and final-outcome labels via the WebRL-trained ORM, and filtered by perplexity to retain only informative trajectories. Failures on the current \emph{training} tasks are analyzed to resample related tasks from a pool of 1{,}573 LLM-generated instructions, with Gemini-2.5-Flash helping assess similarity; the pool is filtered separately for functional overlap with the 165 evaluation tasks (Appendix~\ref{apdx:iterative_rl}); for sparse categories with limited pool coverage (e.g., \emph{Map}), new instructions are synthesized by perturbing failure samples. Over successive phases, this failure-driven loop gradually expands the task distribution while the policy internalizes successful behaviors through potential-shaped offline RL. The full algorithm and implementation details are provided in Appendix~\ref{apdx:iterative_rl}.

 
\section{Experiments}
\label{sec:experiments}
 
\subsection{Experimental Setup}
\label{sec:setup}
 
\paragraph{Benchmarks.}
We evaluate primarily on WebArena-Lite~\citep{liu2024visualagentbench},
a curated and verified subset of WebArena~\citep{zhou2023webarena}
comprising 165 tasks across five domains: \textit{Shopping Admin (CMS)},
\textit{Map}, \textit{Shopping (OSS)}, \textit{Reddit}, and \textit{Gitlab}.
To assess sensitivity to the curated subset, we additionally evaluate
selected Gemma3-12B methods on the broader 812-task WebArena suite.
This is a broader evaluation within the same environment family, \emph{not}
an independent benchmark or evidence of cross-environment generalization.

\paragraph{Baselines.}
For open-source agents, we employ Llama-3.1~(8B) and Gemma-3~(12B) as backbones, comparing against SFT, AWR~\citep{peng2019advantage}, DigiRL~\citep{bai2024digirl}, and WebRL~\citep{qi2024webrl}. For reference, we also report proprietary models GPT-4-Turbo, GPT-4o, Gemini-2.5-Flash, and Gemini-2.5-Pro.\footnote{Although newer Gemini and OpenAI models were released while this study was underway, we could not re-run all experiments with them. We chose to base our evaluation on these models to ensure a fair and comprehensive assessment of all method variants.} All learning-based methods are initialized from the same SFT checkpoints (Llama-3.1-SFT and Gemma-3-SFT). Critic networks share the actor backbone with randomly initialized value heads. For MiRA, the potential critic is pre-trained on an offline dataset before the curriculum commences. We reproduce DigiRL within WebArena-Lite, retaining its AWR optimization and experience replay buffer with 6 rounds of online interaction. For WebRL, we use the official Llama-3.1 checkpoint and train Gemma-3 from scratch following the same curriculum. Static baselines (SFT, AWR) appear as dashed horizontal lines in SR evolution curves. The original WebRL release reports 42.4\% for Llama3.1-8B in its own setup, while the 38.8\% in our table is our unified-harness evaluation of the released checkpoint (Appendix~\ref{sec:apdx_exp}). Our most direct comparison is the budget- and backbone-matched Gemma3 WebRL reproduction (35.1\%) versus MiRA (43.0\%).
 
\subsection{Results: Multi-Phase Training, Domain Trends, and Ablations}
\label{sec:main_results}

\begin{table}[t]
    \centering
    \footnotesize
    \caption{\textbf{Task Success Rate (SR) comparison on WebArena-Lite.} Comparison of MiRA against proprietary and open-sourced baselines. We report Pass@1 to emphasize immediate task completion capability.}
    \label{tab:main_results}
    \renewcommand{\arraystretch}{1.0}
    \setlength{\tabcolsep}{5pt}
    \begin{tabular}{l c c c c c c c}
        \toprule
        \textbf{Models} & \textbf{\#Params} & \textbf{Reddit} & \textbf{Gitlab} & \textbf{CMS} & \textbf{Map} & \textbf{OSS} & \textbf{Avg. SR} \\
        \midrule
        \multicolumn{8}{c}{\textit{Proprietary LLMs}} \\
        \midrule
        GPT-4-Turbo & N/A & 10.5 & 16.7 & 14.3 & 36.7 & 13.3 & 17.6 \\
        GPT-4o & N/A & 10.5 & 10.0 & 20.0 & 20.0 & 11.1 & 13.9 \\
        Gemini-2.5-flash & N/A & 21.1 & 33.3 & 14.3 & 11.5 & 24.4 & 20.6 \\
        Gemini-2.5-pro & N/A & 31.6 & 30.0 & 34.3 & 15.4 & 13.3 & 23.0 \\
        \midrule
        \multicolumn{8}{c}{\textit{Small Open-sourced LLMs}} \\
        \midrule
        AutoWebGLM~\citep{lai2024autowebglm} & 6B & 9.4 & 15.0 & 28.6 & 24.8 & 17.1 & 18.2 \\
        GLM-4-Chat~\citep{glm2024chatglm} & 9B & 5.3 & 10.0 & 6.7 & 3.3 & 6.7 & 6.1 \\
        GLM-4 + SFT (BC) & 9B & 47.4 & 13.3 & 31.4 & 23.3 & 13.3 & 22.4 \\
        Llama3.1 + SFT (BC) & 8B & 36.8 & 6.7 & 20.0 & 33.3 & 17.8 & 20.6 \\
        Llama3.1 + AWR~\citep{peng2019advantage} & 8B & 57.9 & 26.7 & 31.4 & 26.7 & 17.8 & 28.5 \\
        Llama3.1 + DigiRL~\citep{bai2024digirl} & 8B & 57.9 & 26.7 & 37.1 & 33.3 & 17.8 & 30.3 \\
        Llama3.1 + WebRL~\citep{qi2024webrl} & 8B & 68.4 & 36.7 & 51.4 & 15.4 & \textbf{40.0} & 38.8 \\
        \midrule
        Gemma3 + SFT (BC) & 12B & 52.6 & 40.0 & 42.9 & 23.1 & 17.8 & 30.9 \\
        Gemma3 + DigiRL & 12B & 52.6 & 46.7 & 37.1 & \textbf{34.6} & 20.0 & 33.3 \\
        Gemma3 + WebRL & 12B & 68.4 & 43.3 & 40.0 & 30.8 & 20.0 & 35.1 \\
        \rowcolor{red!5}
        \textbf{Gemma3 + MiRA (ours)} & \textbf{12B} & \textbf{73.7} & \textbf{56.7} & \textbf{54.3} & 30.8 & 28.9 & \textbf{43.0} \\
        \bottomrule
    \end{tabular}
\end{table}
 
Table~\ref{tab:main_results} presents the main end-to-end results through the whole fine-tuning effort. Gemma3 + MiRA (12B) achieves the highest average success rate of \textbf{43.0\%}, substantially outperforming the WebRL baseline (35.1\%) and DigiRL (33.3\%) under the same Gemma3-SFT initialization. The margin is largest in complex domains such as \textit{Gitlab} (56.7\%) and \textit{Shopping Admin} (54.3\%), where strict procedural dependencies make sparse-reward methods particularly fragile. MiRA also surpasses all proprietary baselines by a wide margin (e.g., +20.0\% absolute over Gemini-2.5-Pro), demonstrating that milestone-driven shaping closes---and reverses---the gap between open and closed models. To understand how these gains emerge, we next examine the training dynamics across curriculum phases, the per-domain breakdown, and a systematic ablation of each MiRA component.

\paragraph{Additional evaluations.}
On the full 812-task WebArena suite, Gemma3-12B SFT, WebRL, and MiRA obtain
18.8\%, 24.0\%, and \textbf{30.4\%} Pass@1 SR, respectively: a 6.4-point
margin over matched WebRL. On Llama3.1-8B in the Lite evaluation, MiRA reaches
\textbf{41.2\%} ($\pm 1.2$ run-level SE), versus WebRL's 38.8\%
($\pm 1.1$). These results support robustness within the evaluated WebArena
setting and across these two backbones, not transfer to independent environments.

\begin{figure}[t!]
    \centering
    \begin{subfigure}[t]{0.45\linewidth}
        \centering
        \includegraphics[width=\linewidth]{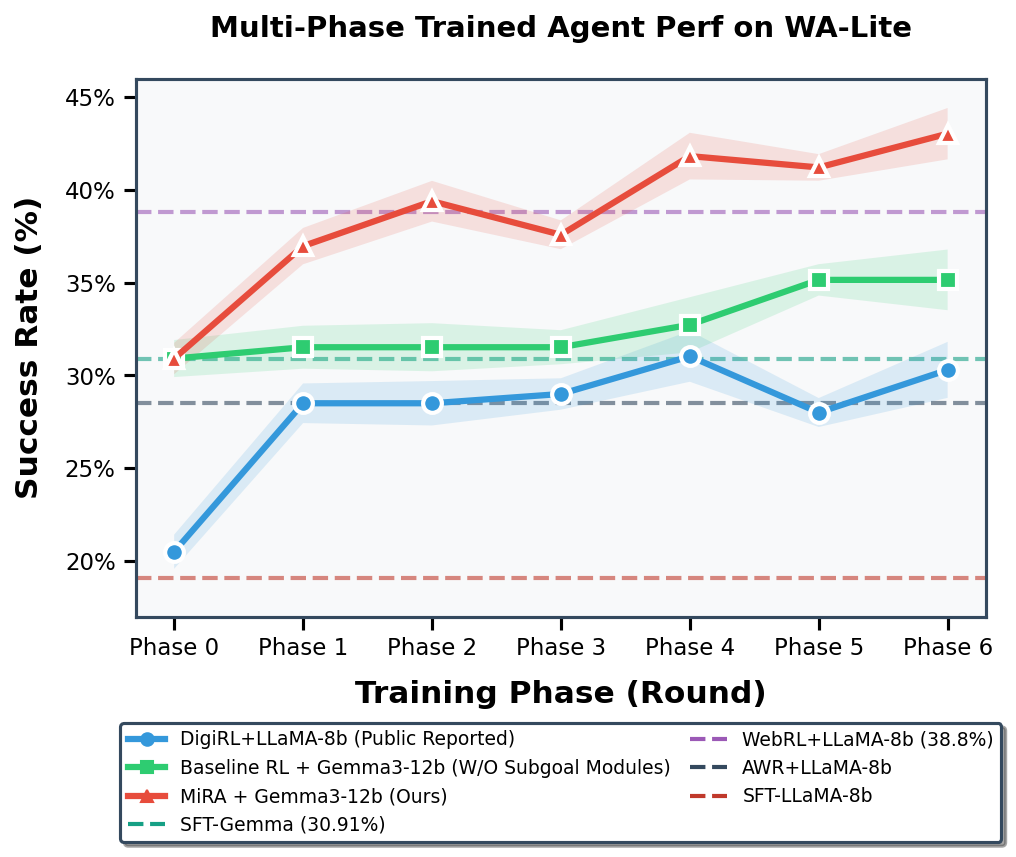}
        \caption{Overall SR across phases.}
        \label{fig:sr_overall}
    \end{subfigure}
    \hfill
    \begin{subfigure}[t]{0.45\linewidth}
        \centering
        \includegraphics[width=\linewidth]{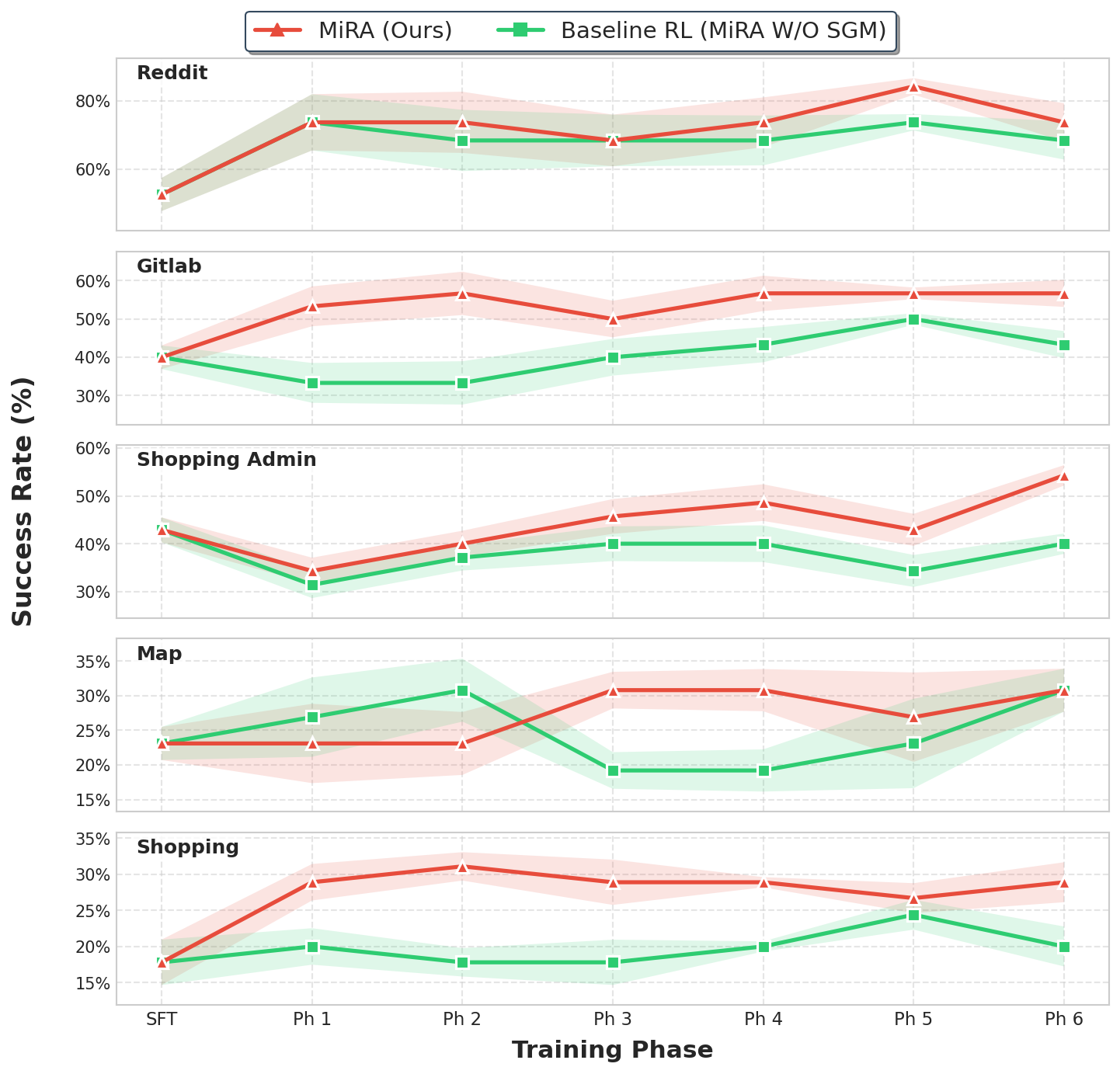}
        \caption{Per-site SR trends.}
        \label{fig:sr_site}
    \end{subfigure}
    \caption{\textbf{Multi-phase and domain-level success rate trends.} \textbf{(a)} MiRA (red) continually improves with each refinement phase, outperforming both standard RL (green) and previous methods such as WebRL and SFT. \textbf{(b)} Per-domain SR gains of MiRA vs.\ baseline RL, consistent across Reddit, Gitlab, CMS, Map, and Shopping (OSS). Results averaged over 5 runs.}
    \label{fig:sr_combined}
    \vspace{-5pt}
\end{figure}

\paragraph{Training dynamics and subgoal chaining.}

\begin{wrapfigure}{r}{0.53\textwidth}
    \centering
    \vspace{-10pt}
    \includegraphics[width=\linewidth]{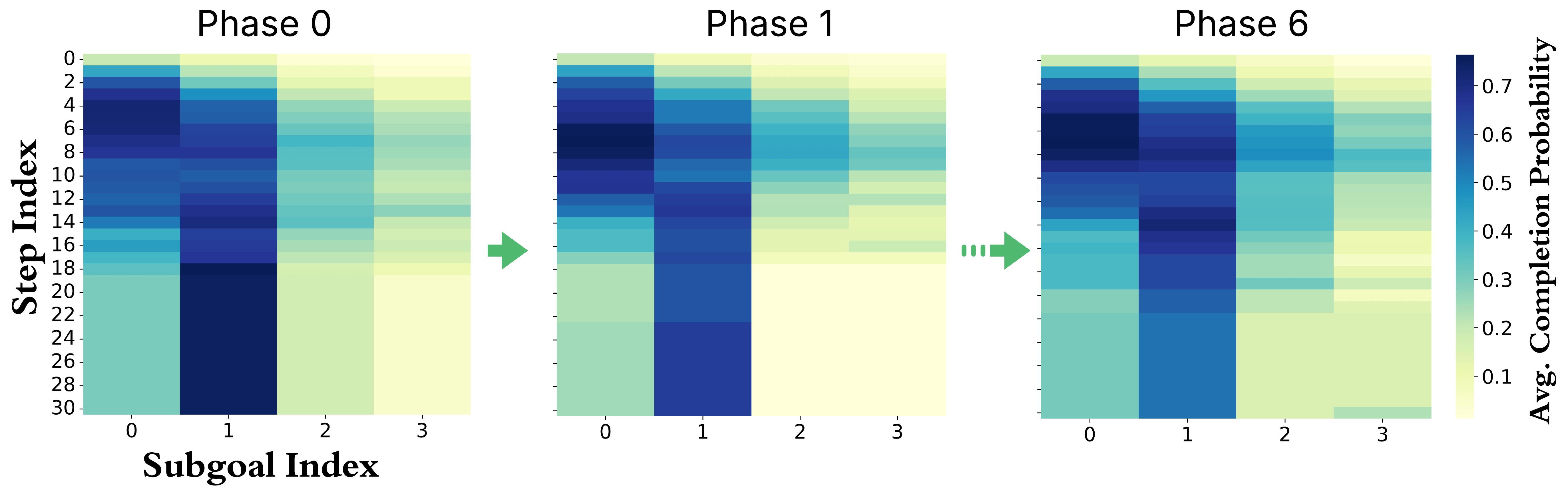}
    \caption{\textbf{Subgoal completion patterns.} Phase~0 stagnates at Subgoal~1; Phase~6 exhibits a diagonal gradient indicating sequential chaining of subgoals over the episode horizon.}
    \label{fig:subgoal_pattern}
    \vspace{-10pt}
\end{wrapfigure}

Across all training rounds, MiRA steadily improves from an initial SR of ${\sim}31\%$ to \textbf{43\%}, while the baseline RL model without subgoal shaping saturates near 35\% (Figure~\ref{fig:sr_combined}a).  Per-domain analysis (Figure~\ref{fig:sr_combined}b) shows consistent upward trends, particularly on Gitlab and Shopping Admin, where subgoal-based shaping provides denser intermediate rewards for complex multi-step interactions.

Figure~\ref{fig:subgoal_pattern} visualizes the average subgoal completion probability over the episode horizon. In Phase~0, probability mass concentrates as a vertical band over the first two subgoals---the agent initiates tasks but fails to transition to intermediate objectives. By Phase~6, the distribution shifts to a structured \emph{diagonal gradient}, demonstrating that MiRA has enabled \emph{sequential fluency}: the agent efficiently chains subgoals in lockstep with the episode timeline, transforming fragmented attempts into coherent, long-horizon trajectories.

\begin{figure}[t]
    \centering
    \begin{subfigure}[b]{0.43\textwidth}
        \centering
        \includegraphics[width=\linewidth]{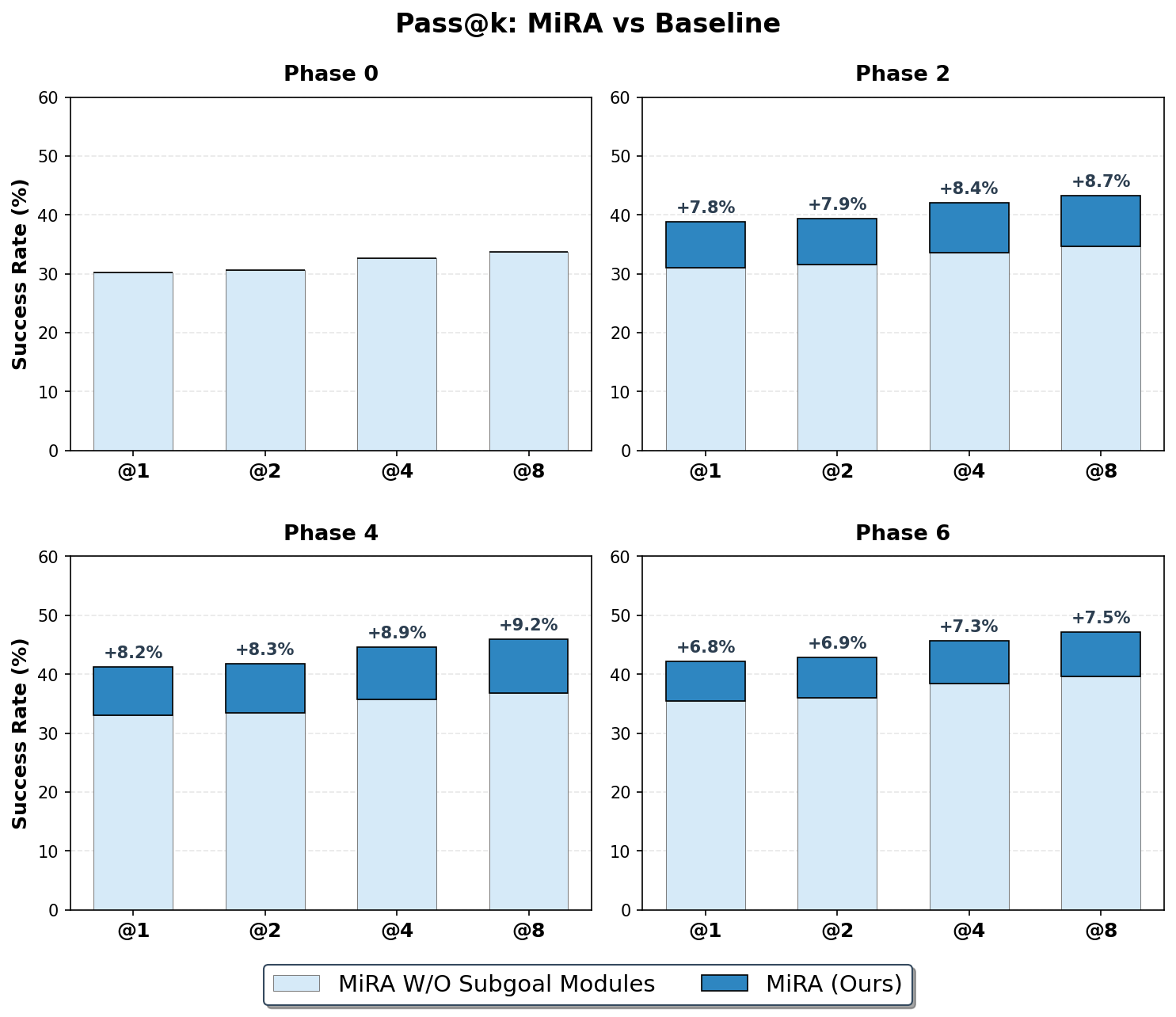}
        \caption{\textbf{Pass@$k$ scaling.} Performance across varying sample budgets ($k \in \{1,2,4,8\}$) at different training stages.}
        \label{fig:pass_k_analysis}
    \end{subfigure}
    \hfill
    \begin{subfigure}[b]{0.43\textwidth}
        \centering
        \includegraphics[width=\linewidth]{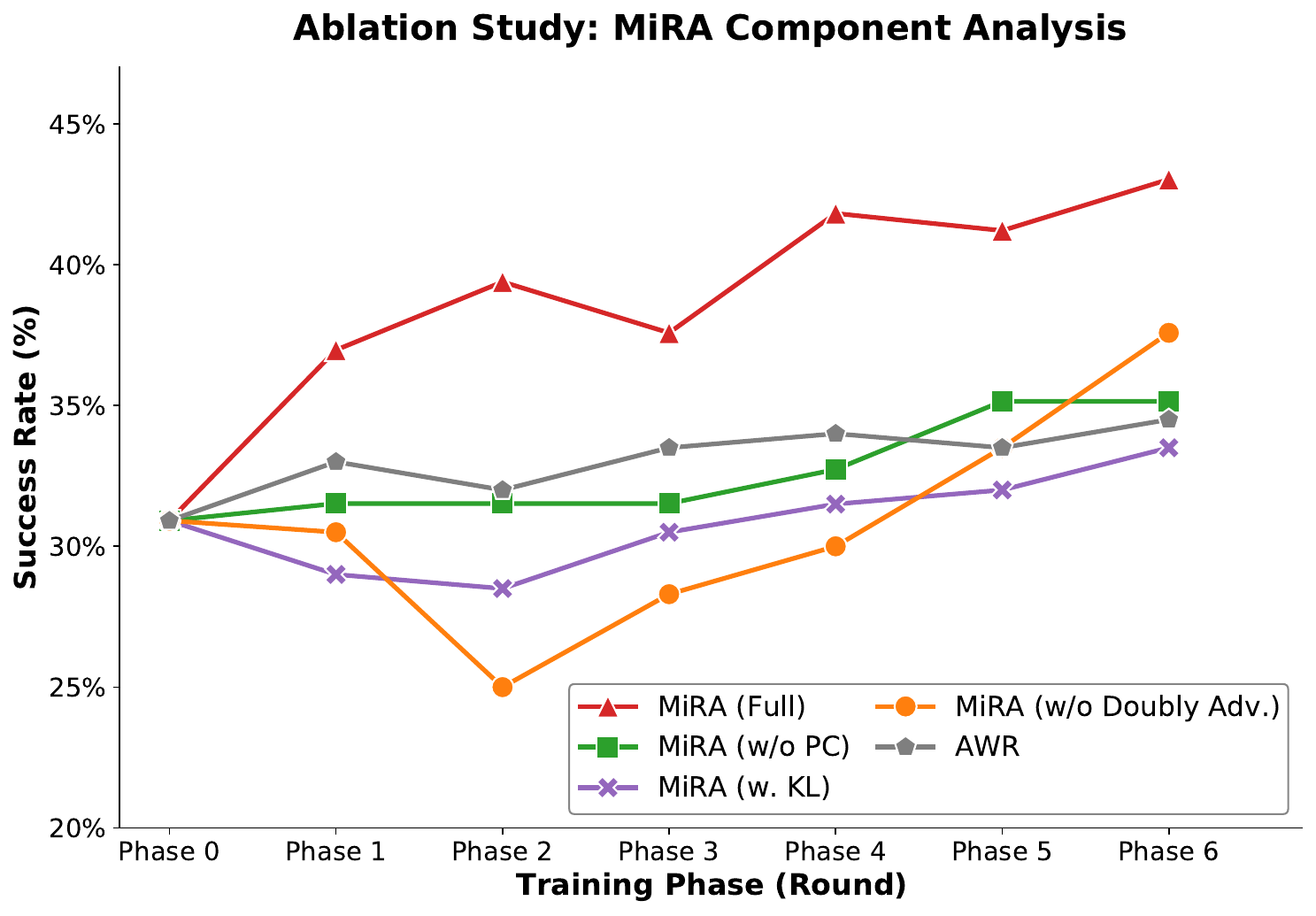}
        \caption{\textbf{Component ablation.} Impact of removing the potential critic (PC), mixed TD--Monte-Carlo estimation, and MSE updates. Each curve averages 3 trials from the same SFT init ($\text{SR}{\approx}31\%$).}
        \label{fig:ablation_components}
        
    \end{subfigure}
    \caption{\textbf{Robustness and component analysis.} \textbf{(a)} MiRA consistently outperforms the baseline across all sample budgets ($+7.9\%$ at pass@2 in Phase~2; $+7.5\%$ at pass@8 in Phase~6). \textbf{(b)} Removing any key component degrades SR toward the SFT baseline, validating the full framework.}
    \label{fig:combined_analysis}
    \vspace{-5pt}
\end{figure}

\paragraph{Pass@$k$ consistency and ablated variants.}
For a task with $c$ successes out of $n$ rollouts, we compute pass@$k$ for $k\in\{1,2,4,8\}$ using the unbiased estimator $\text{pass}@k = 1 - \binom{n-c}{k}/\binom{n}{k}$. As shown in Figure~\ref{fig:pass_k_analysis}, MiRA consistently outperforms the baseline across all sample budgets; the gap widens in Phase~2 ($+7.9\%$ at pass@2) and MiRA maintains a strong lead through Phase~6 ($+7.5\%$ at pass@8). To probe whether simpler alternatives suffice, we compare five configurations: (1)~\textbf{MiRA~(Full)}: the complete framework with MSE regression (Eq.~\ref{eq:mse_objective}), potential critic, and mixed TD--Monte-Carlo target (Eq.~\ref{eq:advantage}); (2)~\textbf{MiRA~w/o~PC}: no potential critic, sparse outcome rewards only; (3)~\textbf{MiRA~w.~KL}: replaces MSE with direct KL minimization; (4)~\textbf{MiRA~w/o~Doubly~Adv.}: only 1-step TD error ($\lambda{=}1$); (5)~\textbf{AWR}~\citep{peng2019advantage}: a simpler off-policy baseline.

\paragraph{Component analysis.}
Each ablated variant isolates a distinct failure mode visible in Figure~\ref{fig:ablation_components}. \textbf{(i) Mixed TD--Monte-Carlo target:} Removing MC mixing (orange) causes performance to \emph{collapse} to ${\approx}25\%$ in early phases because the under-calibrated $V_\phi$ propagates bias directly into advantage estimates via the 1-step TD error. By Phase~6 the variant recovers to ${\approx}37\%$, but the early damage delays convergence; the MC return $G_t$ reduces dependence on the early critic, although replay-based estimates are not necessarily unbiased. \textbf{(ii) Potential critic:} Without $P_\psi$ (green), learning plateaus at ${\approx}35\%$---the agent cannot credit intermediate actions for eventual success, confirming subgoal-based potential shaping is essential for long-horizon credit assignment. The KL variant (purple) also underperforms (${\approx}33\%$ by Phase~6); a full justification for MSE over KL divergence is provided in Appendix~\ref{sec:apdx_mse_vs_kl}. Together, these components address complementary failure modes, and removing any single one results in substantial degradation. Additional parameter-sensitivity ablations are in Appendix~\ref{sec:apdx_ablation}.

\section{Conclusion}
\label{sec:discussion}
 
We have presented MiRA, a framework that enhances long-horizon planning in web agents through milestone-based potential shaping. By combining an automated failure analysis with validated subgoal generation and a dual-critic RL architecture, MiRA boosts Gemma3-12B to 43.0\% on WebArena-Lite---surpassing both proprietary systems and prior open-source state of the art. Our evidence identifies sparse-reward credit assignment as an important bottleneck and shows that semantically grounded milestones provide useful auxiliary training feedback in the evaluated WebArena settings. This is an empirical result rather than a policy-invariance or general cross-environment guarantee. Extended discussion connecting MiRA to process reward models, the subgoal grounding trade-off, and the path toward self-evolving autonomous agents is provided in Appendix~\ref{sec:apdx_discussion}.

\section*{Acknowledgments and Disclosures}

\textbf{Acknowledgments.}
This work was conducted while Taiyi Wang was a member of the Frontier AI organization at Google DeepMind. We gratefully acknowledge our colleagues in the Frontier AI team for their support, insightful discussions, and valuable feedback throughout this research.

\textbf{Funding and support.}
This work was supported by Google DeepMind through institutional research resources.

\textbf{Competing interests.}
The authors declare no competing interests.


\bibliographystyle{abbrvnat}

\bibliography{main}

\newpage

\appendix

\section{Appendix}

\subsection{Proof of PBRS Equivalence for Goal-Conditioned Tasks}
\label{sec:apdx_proof}

We provide the mathematical proof that Potential-Based Reward Shaping (PBRS) is equivalent to training with the original reward function but initializing the Q-function to the potential function $\Phi$. This proof holds for the goal-conditioned setting under the standard PBRS assumptions, including compatible discounting and terminal potentials, where the reward, value, and potential functions depend on a goal $g \in \mathcal{G}$. \textbf{It does not prove policy invariance for MiRA's implemented shaping or actor target.}

\subsubsection{Equivalence of Bellman Updates}

Let's define two learning scenarios for a goal-conditioned Bellman optimality update.

\begin{itemize}
    \item 
    \textbf{Scenario A: PBRS (Shaped Reward)}
    \begin{itemize}
        \item Reward: $\tilde{r}(s, a, s', g) = r(s, a, s', g) + \gamma \Phi(s', g) - \Phi(s, g)$
        \item Q-function: $\tilde{Q}_k(s, a, g)$
        \item Initialization: $\tilde{Q}_0(s, a, g) = 0$
        \item Bellman Update: $\tilde{Q}_{k+1}(s, a, g) = \mathbb{E}_{s' \sim P(\cdot|s,a)} \left[ \tilde{r}(s, a, s', g) + \gamma \max_{a'} \tilde{Q}_k(s', a', g) \right]$
    \end{itemize}
    
    \item
    \textbf{Scenario B: $\Phi$-Initialization (Original Reward)}
    \begin{itemize}
        \item Reward: $r(s, a, s', g)$
        \item Q-function: $Q_k(s, a, g)$
        \item Initialization: $Q_0(s, a, g) = \Phi(s, g)$
        \item Bellman Update: $Q_{k+1}(s, a, g) = \mathbb{E}_{s' \sim P(\cdot|s,a)} \left[ r(s, a, s', g) + \gamma \max_{a'} Q_k(s', a', g) \right]$
    \end{itemize}
\end{itemize}

\begin{claim}[Wiewiora, 2003, adapted for goal-conditioning]
For any goal $g \in \mathcal{G}$ and all $k \ge 0$, the Q-functions from the two scenarios are related by:
$$ \tilde{Q}_k(s, a, g) = Q_k(s, a, g) - \Phi(s, g) $$
\end{claim}

\begin{proof}[Proof by Induction]
We prove this claim by induction on the iteration step $k$.

\textbf{Base Case (k=0):}
We check the relationship for the initial Q-values:
\begin{align*}
    \tilde{Q}_0(s, a, g) &= Q_0(s, a, g) - \Phi(s, g) \\
    0 &= \Phi(s, g) - \Phi(s, g) \\
    0 &= 0
\end{align*}
The base case holds.

\textbf{Inductive Step:}
Assume the relationship holds for an arbitrary step $k$. This is our inductive hypothesis:
$$ \tilde{Q}_k(s, a, g) = Q_k(s, a, g) - \Phi(s, g) $$
We must now prove that it also holds for $k+1$. We start with the Bellman update for Scenario A:
$$ \tilde{Q}_{k+1}(s, a, g) = \mathbb{E}_{s' \sim P(\cdot|s,a)} \left[ \tilde{r}(s, a, s', g) + \gamma \max_{a'} \tilde{Q}_k(s', a', g) \right] $$
Substitute the PBRS reward definition $\tilde{r}$:
$$ \tilde{Q}_{k+1}(s, a, g) = \mathbb{E}_{s'} \left[ (r(s, a, s', g) + \gamma \Phi(s', g) - \Phi(s, g)) + \gamma \max_{a'} \tilde{Q}_k(s', a', g) \right] $$
Now, substitute the inductive hypothesis $\tilde{Q}_k(s', a', g) = Q_k(s', a', g) - \Phi(s', g)$ into the $\max$ term:
$$ \tilde{Q}_{k+1}(s, a, g) = \mathbb{E}_{s'} \left[ (r + \gamma \Phi(s', g) - \Phi(s, g)) + \gamma \max_{a'} \left( Q_k(s', a', g) - \Phi(s', g) \right) \right] $$
The potential $\Phi(s', g)$ is a function of state $s'$ and goal $g$, but not action $a'$. Therefore, it is a constant with respect to the $\arg\max$ operation over $a'$ and can be pulled out:
$$ \max_{a'} \left( Q_k(s', a', g) - \Phi(s', g) \right) = \left( \max_{a'} Q_k(s', a', g) \right) - \Phi(s', g) $$
Substitute this back into our main equation:
\begin{align*}
    \tilde{Q}_{k+1}(s, a, g) &= \mathbb{E}_{s'} \left[ (r + \gamma \Phi(s', g) - \Phi(s, g)) + \gamma \left( \left( \max_{a'} Q_k(s', a', g) \right) - \Phi(s', g) \right) \right] \\
    &= \mathbb{E}_{s'} \left[ r + \gamma \Phi(s', g) - \Phi(s, g) + \gamma \max_{a'} Q_k(s', a', g) - \gamma \Phi(s', g) \right]
\end{align*}
The $\gamma \Phi(s', g)$ terms cancel out:
$$ \tilde{Q}_{k+1}(s, a, g) = \mathbb{E}_{s'} \left[ r(s, a, s', g) + \gamma \max_{a'} Q_k(s', a', g) - \Phi(s, g) \right] $$
Since $\Phi(s, g)$ depends on the current state $s$ and goal $g$, but not on the next state $s'$, we can pull it out of the expectation over $s'$:
$$ \tilde{Q}_{k+1}(s, a, g) = \left( \mathbb{E}_{s'} \left[ r(s, a, s', g) + \gamma \max_{a'} Q_k(s', a', g) \right] \right) - \Phi(s, g) $$
The term in the parentheses is exactly the Bellman update for $Q_{k+1}(s, a, g)$ from Scenario B.
$$ \tilde{Q}_{k+1}(s, a, g) = Q_{k+1}(s, a, g) - \Phi(s, g) $$
This completes the inductive step.
\end{proof}

\subsubsection{Implication: Goal-Conditioned Policy Invariance}

The result $\tilde{Q}_k(s, a, g) = Q_k(s, a, g) - \Phi(s, g)$ proves that the goal-conditioned Q-function learned with PBRS is simply the ``true'' goal-conditioned Q-function offset by the potential $\Phi(s, g)$. This directly proves policy invariance for any given goal $g$.

A greedy, goal-conditioned policy $\pi_k$ at step $k$ is derived by:
$$ \pi_k(s, g) = \arg\max_{a} Q_k(s, a, g) $$
If we derive the policy $\tilde{\pi}_k$ from the PBRS Q-function $\tilde{Q}_k$:
$$ \tilde{\pi}_k(s, g) = \arg\max_{a} \tilde{Q}_k(s, a, g) $$
Substitute our proven relationship:
$$ \tilde{\pi}_k(s, g) = \arg\max_{a} \left( Q_k(s, a, g) - \Phi(s, g) \right) $$
Since $\Phi(s, g)$ is a constant with respect to the $\arg\max$ over $a$ (as it does not depend on the action $a$), it does not change which action yields the maximum value.
$$ \arg\max_{a} \left( Q_k(s, a, g) - \Phi(s, g) \right) = \arg\max_{a} Q_k(s, a, g) $$
Therefore, for any state $s$ and goal $g$:
$$ \tilde{\pi}_k(s, g) = \pi_k(s, g) $$
The policy derived from the shaped Q-function is identical to the policy derived from the original Q-function at every step $k$. Under the conditions of the \emph{classical} PBRS rule above, the optimal action selection is unchanged. This statement is not a guarantee for MiRA.

The shaping-weight selection and its limitations are reported in Appendix~\ref{sec:apdx_ablation}.

\paragraph{Interpretation of MiRA's actor target.}
Let $\Delta P_t=P_\psi(s_{t+1},g)-P_\psi(s_t,g)$ and
$G_t^{\mathrm{out}}=\sum_{u=t}^{T}\gamma^{u-t}r_u$.
Expanding Eq.~\ref{eq:advantage} gives the algebraic decomposition
\begin{align}
A_t^{\mathrm{shaped}} &= A_t^{\mathrm{out}}
+\alpha\left[\lambda\Delta P_t+(1-\lambda)
\sum_{u=t}^{T}\gamma^{u-t}\Delta P_u\right],\\
A_t^{\mathrm{out}} &= \lambda\bigl(r_t+\gamma V_\phi(s_{t+1},g)
-V_\phi(s_t,g)\bigr)
+(1-\lambda)\bigl(G_t^{\mathrm{out}}-V_\phi(s_t,g)\bigr).
\end{align}
Thus the actor regresses onto an outcome-anchored quantity plus a controlled
progress term. Because $V_\phi$ is trained only on terminal outcomes while
$G_t$ contains shaped rewards, this quantity is not an unbiased advantage for
either the original MDP or a conventional PBRS-shaped MDP. Its use is an
empirically evaluated surrogate, not a proof of preserved optimal policies.

\subsection{Detailed Failure Analysis}
\label{sec:apdx_failure}

This appendix provides the full failure analysis content that motivated our framework, building on the condensed motivation in the main text. We developed a comprehensive automated analyzer that performs large-scale trajectory evaluation and categorization. Leveraging the reasoning and perception capabilities of the Gemini-2.5-Flash model, our analyzer systematically inspects and summarizes tons of recorded trajectories to uncover behavioral patterns. In short, our analyzer performs three core functions: \textbf{(1)} it first acts as an objective judge, providing outcome-based summarization; \textbf{(2)} it then performs failure categorization into a specific error mode; and \textbf{(3)} it conducts a root-cause analysis to pinpoint the exact decision step where the agent deviated.

\subsubsection{Function 1: Trajectory Evaluation and Summarization}

As its first function, the analyzer provides an objective \textbf{Failure Summarization}. This process determines \textit{if} a failure occurred based on hardcoded behavioral rules before passing the trajectory to the next stage of diagnosis. This procedure gives out the failure summary: A trajectory is marked as a failure solely based on the outcome of the final benchmark checking rules (i.e., \emph{``Only judge, never guess''}). This ensures that classification is objective and free of bizarre hallucinations, focusing mostly on the key state validation (e.g., Achieved XX, not achieved XX). A typical objective summary, adhering to this rule, would be:
    \begin{tcolorbox}[
        colback=blue!5!white,
        colframe=blue!75!black,
        title=\textbf{Failure Summary},
        fonttitle=\bfseries\texttt,
        colbacktitle=blue!20!white,
        coltitle=black,
        rounded corners,
        arc=4pt,
        boxrule=1pt
    ]
    \texttt{``Agent navigated to Hollidaysburg (film) page. Agent extracted information from this page. Agent failed to move forward and terminated. Final benchmark check: FAILED.''}
    \end{tcolorbox}

\subsubsection{Function 2: Categorization via Prioritized Hardcoded Rules}

Once a trajectory is confirmed as a failure by the first function, the analyzer's second function is \textbf{Categorization via Prioritized Hardcoded Rules}. Failed trajectories are systematically classified into one of four mutually exclusive failure modes using the hard-coded rules detailed in Table~\ref{tab:failure_rules}, which are applied in a strict, descending order of priority.

\begin{table}[htbp]
    \centering
    \caption{\textbf{Hardcoded Rules for Failure Mode Classification}}
    \label{tab:failure_rules}
    \begin{tabularx}{\textwidth}{>{\bfseries}p{4cm} | >{}X} 
        \toprule
        Category & Hardcoded Rules \\
        \midrule
        Stop at Wrong Page (\textit{Wrong Termination}) & The agent called the $\text{exit()}$ function or terminated, but the message was judged \textit{incorrect} by the final benchmark rules. \\
        \midrule
        Get Stuck Midway & The agent failed and exhibited repetition, characterized by identical last $N$ actions or a short action-state sequence repeating $M$ times. \\
        \midrule
        Fail to Make Reasonable Attempt & The agent already reached the right page, but didn't attempt to $\text{exit()}$ \textbf{OR} The agent failed very early, either by an immediate, short $\text{exit()}$ call or by clicking a nonsensical, distracting element as its first action. \\
        \midrule
        Others (\textit{System error, etc.}) & The agent failed, but did not trigger Rules A, B, or C, indicating a general deviation from the success path. \\
        \bottomrule
    \end{tabularx}
\end{table}

\subsubsection{Function 3: Identifying the Decision Step}

\begin{wrapfigure}{r}{0.6\textwidth}
    \centering
    \vspace{-10pt}
    \includegraphics[width=0.58\textwidth]{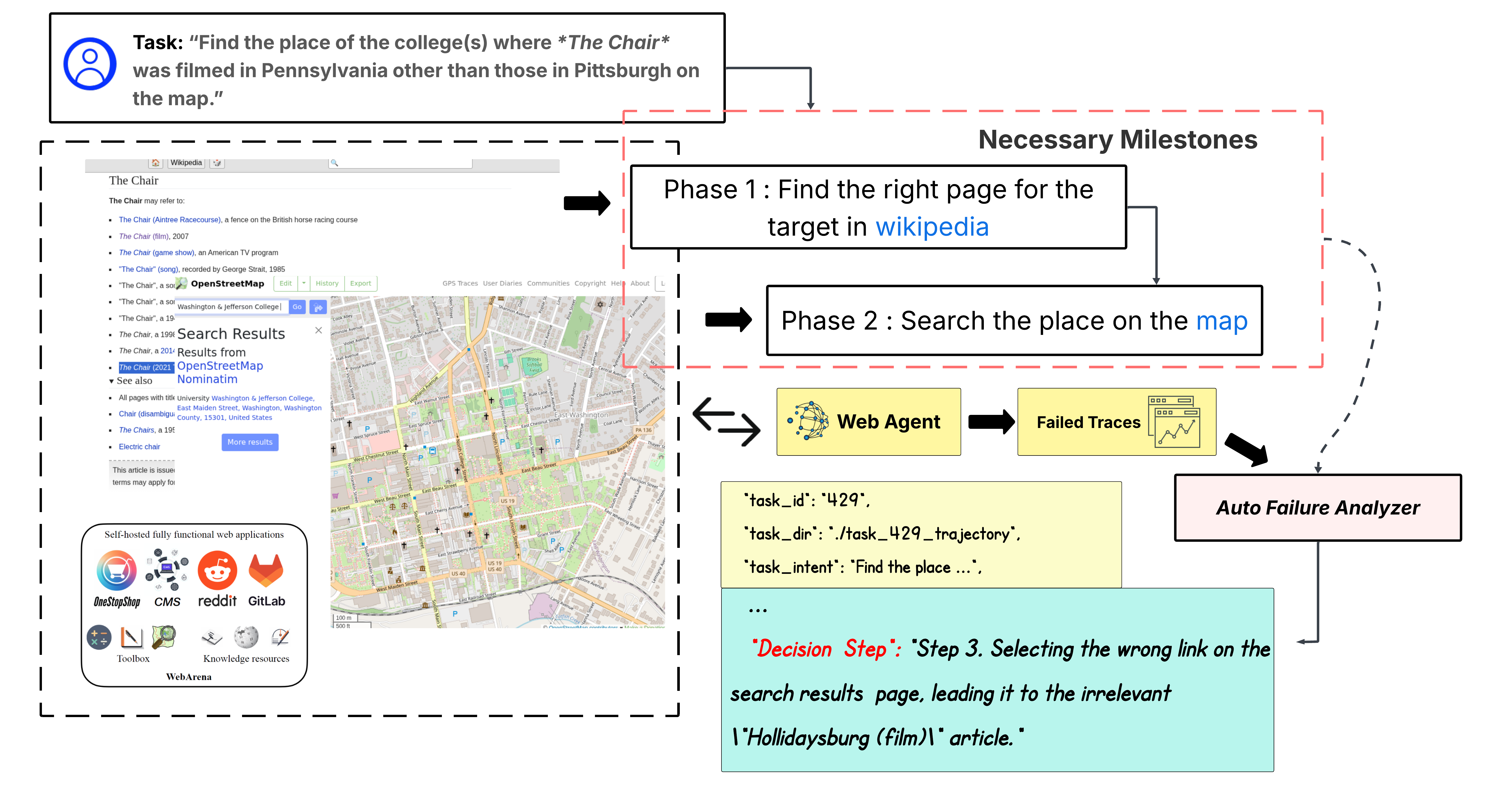}
    \caption{\textbf{Automated Identification of Key Decision Steps.} The analyzer compares the agent's trajectory against task intent and known successful patterns to pinpoint the exact step (e.g., Step 3) where a critical deviation occurred.}
    \label{fig:decision_step_example}
    \vspace{-10pt}
\end{wrapfigure}

The third and final function of the analyzer is to pinpoint the exact \textit{Key Decision Step} where the agent deviated from an optimal or successful trajectory. This capability moves beyond simple error categorization to actionable root-cause analysis. To automate this pinpointing process with high fidelity, we employ a differential analysis strategy. The analyzer aligns the agent's failed trajectory $\tau_{\text{fail}}$ against a set of \textit{Reference Success Traces} derived from two distinct sources:
 \textbf{(1) Teacher Demonstrations:} Successful trajectories generated by superior, proprietary reasoning models (e.g., IBM-CUGA) which serve as an upper-bound oracle for reasoning quality. \textbf{(2) Golden Paths from Peers:} Validated ground-truth sequences from the training dataset or historical successful runs by other trained agents (e.g., WebRL agent). Then the analyzer scans for the \textbf{First Point of Significant Divergence}---the earliest timestep $t$ where the agent's action $a_t$ semantically contradicts the reference strategy. This method isolates the root cause by filtering out downstream errors that are merely symptoms of the initial mistake. Specifically, the system targets two primary divergence patterns:
\begin{itemize}
    \item \textbf{Semantic Deviation:} Identifying the step where the agent chooses a link or search query that conceptually drifts from the cluster of actions found in successful traces (as seen in Step 3 of Figure~\ref{fig:decision_step_example}).
    \item \textbf{Stagnation and Loop Onset:} For ``Get Stuck'' failures, the analyzer utilizes state-hashing to detect cyclic sub-graphs in the interaction history. It retroactively marks the entry point into the loop—rather than the loop itself—as the key decision step, highlighting the moment the agent failed to recognize the lack of state progression.
\end{itemize}

As illustrated in Figure~\ref{fig:decision_step_example}, the analyzer parses the agent's complete trajectory against the task intent. In this typical example (Task ID 429), the goal is to find a specific filming location for ``The Chair'' in Pennsylvania. The optimal path involves two main milestones (subgoals): (1) finding the correct Wikipedia page to identify the college, and (2) searching for that college on the map.

The auto-analyzer successfully identifies that the critical failure occurred at \textbf{Step 3}, where the agent selected an incorrect link (``The Chair (2007 film)'') instead of identifying the correct TV show or other relevant entries. This single decision cascaded into a failure, as it led the agent down an irrelevant path of information.

The reliability of this identification stems directly from the robust classification rules defined previously (Function 2). Because the analyzer first rigorously classifies the \textit{type} of failure (e.g., ``Get Stuck Midway'' vs. ``Fail to Make Reasonable Attempt''), it can narrow down the search space for the decisive error. For instance, in a ``Get Stuck Midway'' scenario, the decision step is often the entry point into the repetitive loop. In a ``Wrong Termination'' scenario, it is the final action before incorrectly claiming success. By grounding the search for the key decision step in these validated failure modes, the analyzer achieves high precision in identifying the exact moment the agent's reasoning faltered.

\subsubsection{Soundness of the Classifier}

To ensure the reliability of our automated failure categorization, we implemented a rigorous prompting and validation framework.

\begin{wraptable}{r}{0.35\textwidth}
    \centering
    \vspace{-5pt}
    \captionsetup{font=footnotesize}
    \caption{Agreement between automated analyzer and human labels ($N=40$).}
    \label{tab:validation_scores}
    \small
    \begin{tabular}{lr}
        \toprule
        \textbf{Failure Type} & \textbf{Score} \\
        \midrule
        Stuck in Midway & 10/10 \\
        Wrong Termination & 10/10 \\
        Fail Attempt & 8/10 \\
        Others & 9/10 \\
        \bottomrule
    \end{tabular}
    \vspace{-8pt}
\end{wraptable}

\noindent \textbf{Few-Shot Prompts:} The analysis system was primed using four distinct failure examples—one for each defined category—to ground the analyzer's reasoning. This few-shot context provides the Large Language Model with concrete reference points for distinguishing between subtle failure modes (e.g., distinguishing a ``stuck'' agent from one that simply fails to attempt the task).

\noindent \textbf{Validation:} We conducted sanity checks on the classification accuracy using a manually labeled dataset of 40 examples (10 instances per failure type). As shown in Table~\ref{tab:validation_scores}, the analyzer demonstrates high agreement with human annotators, particularly in detecting procedural failures like ``Stuck in Midway'' and ``Wrong Termination'' ($100\%$ accuracy). Slight deviations occurred in the ``Fail to Attempt'' category (8/10), where the boundary between a very short unsuccessful attempt and zero attempt can be semantically ambiguous. Validation also included an analysis of action distributions for each failure type to confirm behavioral consistency across the classified subsets.

\subsubsection{Failure Number Analysis: The Mid-Stuck Challenge}

Quantitative analysis of the failure distribution for various agents (Gemini-2.5-pro, Gemma, and Gemma-SFT) demonstrates that \textbf{Get Stuck Midway} is the most significant challenge across all tested models.

\begin{figure}[h]
    \centering
    \includegraphics[width=1.0\linewidth]{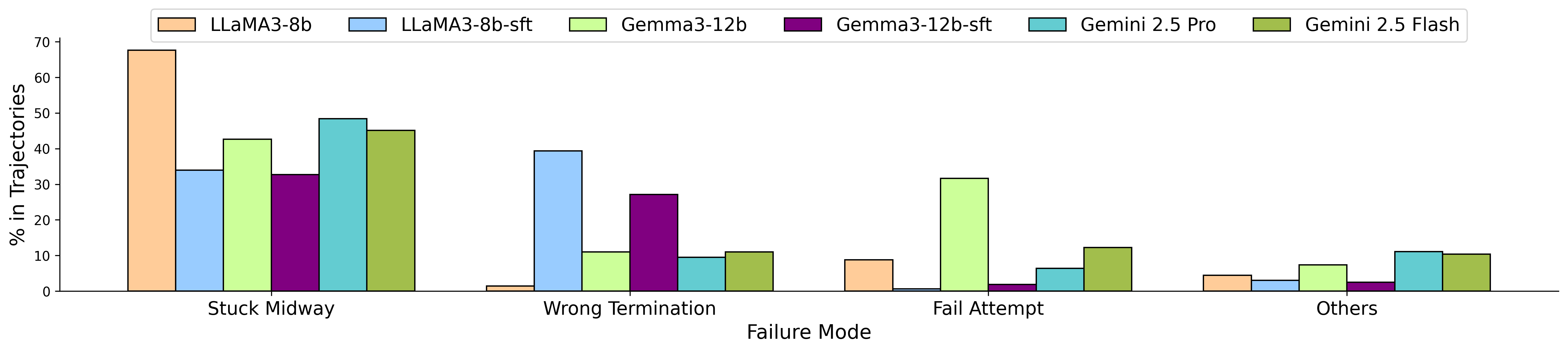}
    \captionsetup{justification=centering}
    \caption{Overview of Failure mode distribution of existing out-of-box models}
    \label{fig:dist}
\end{figure}

\begin{itemize}
    \item \textbf{Midway Dominance:} The majority of failures across all baseline models (Gemini, Gemma) stem from the \emph{Get Stuck Midway} mode, with failure ratios ranging from approximately $42\%$ to $49\%$. This mode is indicative of policy oscillation and poor long-term planning, particularly when the agent encounters unexpected states or is required to execute a long, complex sequence of actions.
    
    \item \textbf{Planning Deficiencies:} This high frequency suggests that existing policies, largely based on Supervised Fine-Tuning (SFT) or similar techniques, lack the robust planning and self-correction mechanisms necessary for sequential, multi-step web tasks. An agent stuck in a loop is fundamentally failing to update its internal state or subgoal, thereby repeating failed actions.
    
    \item \textbf{The Value of Subgoals and Planning:} The other primary failure mode, \emph{Fail to Make Reasonable Attempt}, while significant for Gemma (approximately $32\%$), is less dominant than the Midway failure. Agents that fail midway are often closer to the solution but lack the means to complete the final steps or break an incorrect loop. Therefore, policies integrated with explicit \emph{subgoal completion} or \emph{trajectory planning} signals are expected to yield the largest performance gains by directly targeting the pervasive \emph{Get Stuck Midway} category.
\end{itemize}

\subsection{Resolving the Planning Bottleneck: Post-Training Failure Distribution}
\label{sec:apdx_post_failure}

\begin{figure}[h]
    \centering
    \includegraphics[width=1.0\linewidth]{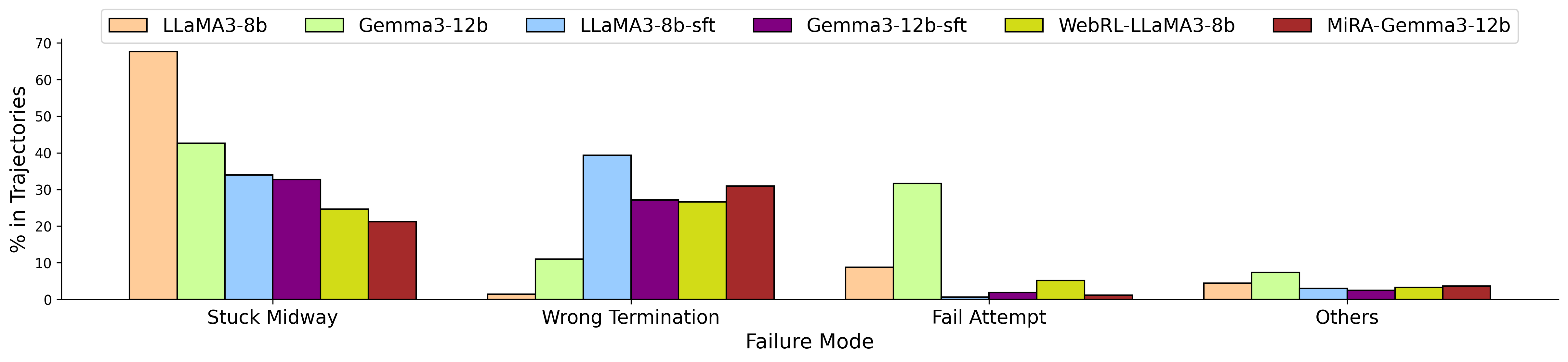}
    \caption{\textbf{Shift in Failure Distribution Post-Training.} Comparison of failure modes across Base, SFT, and RL-tuned models. MiRA (dark red) significantly reduces the ``Stuck Midway'' error rate compared to SFT and standard WebRL baselines, indicating that the potential-based curriculum successfully helps the agent break out of local optima and navigation loops.}
    \label{fig:mira_failure_dist}
\end{figure}

To quantify the impact of MiRA on the agent's fundamental planning capabilities, we applied our automated failure analyzer to the post-training trajectories. Figure~\ref{fig:mira_failure_dist} presents the comparative distribution of failure modes across base LLMs, SFT baselines, the \textsc{WebRL} baseline, and our MiRA agent.

The most significant finding is the drastic reduction in \emph{Get Stuck Midway} errors. As shown in Figure~\ref{fig:mira_failure_dist}, base models (e.g., LLaMA3-8b) and even their SFT counterparts suffer heavily from navigation loops, with Gemma3-12b-SFT exhibiting a stuck rate of $\approx 33\%$. In contrast, MiRA (Gemma3-12B) reduces this category to $\approx 21\%$, a $12$-point reduction over its own Gemma3-SFT initialization. We note that the \textsc{WebRL}-LLaMA3-8B reference ($\approx 25\%$ stuck rate) uses a different backbone and is included as the prior open-source state of the art rather than a matched-backbone control; the SFT-vs-MiRA comparison on the same Gemma3-12B backbone provides the direct evidence that potential-based shaping resolves the planning bottleneck. This reduction serves as strong empirical evidence that our potential-based reward shaping and curriculum strategy effectively smoothen the optimization landscape. By providing dense subgoal signals, MiRA enables the agent to recover from repetitive states where standard policies typically stall, effectively ``pushing'' the agent through long-horizon dependencies.

Interestingly, while MiRA minimizes navigation stalls, we observe a relative increase in \emph{Wrong Termination} errors (rising to $\approx 31\%$). We note that this phenomenon is consistent across all trained agents: since successful termination sequences are highly rewarded during training, optimized policies naturally develop a bias toward attempting termination actions even without full confidence. Furthermore, the rate of wrong termination is compounded by the AutoRater (LLM-as-Judge), which judges success based on reaching the correct terminal page, often failing to verify the semantic details of the final answer. This shift, however, represents a clear progression in capability. Rather than failing to navigate (execution failure), the agent now traverses the full horizon to reach a terminal state, indicating that MiRA has successfully solved the lower-level planning bottleneck. This effectively exposes the higher-level semantic reasoning limitations of the underlying LLM—a preferable trade-off for scalable web automation compared to navigation stagnation.

\subsection{Iterative Policy Refinement and Curriculum Generation}
\label{apdx:iterative_rl}

This section provides additional implementation details for the iterative refinement framework described in Section~\ref{sec:main_results}. The full training pipeline alternates between environment interaction, offline RL optimization, and curriculum generation to progressively improve the agent.

\paragraph{Task-generation quality control and leakage audit.}
Candidates originate from website/domain seeds and controlled task perturbations.
The pipeline applies structural and semantic deduplication, feasibility and
task-specific grading, teacher best-of-$N$ execution checks, pilot-RL/replay
filtering, and human review of representative tasks and failures. Curriculum
evolution is driven by failures in the \emph{training} distribution, not
by evaluating the held-out benchmark. For the 1{,}573 generated-task pool
against the 165 WebArena-Lite evaluation tasks, a hierarchical audit compares
normalized instructions and metadata, intent templates and instantiated
arguments, embedding-based near-duplicate candidates, and finally functional
equivalence (initial state, intended operations, and terminal condition).
After filtering and adjudication, no retained normalized exact duplicates or
task pairs above the configured semantic threshold remain. This reduces
known overlap, but cannot rule out every possible functional similarity.

\paragraph{Matched six-phase budget and curriculum novelty.}
We compare all RL methods under six matched interaction phases and optimization
schedules; six phases are not claimed to prove convergence. In an audit of
200 selected tasks per phase, novelty relative to earlier phases falls from
74\% in Phase~1 to 8\% in Phase~6, with near-duplicate rates rising from
8\% to 68\%. Novelty here uses maximum cross-phase embedding similarity
below 0.85, and near-duplicates exceed 0.92. The number of newly identified
error clusters falls from 31 to 2. These diagnostics motivate stopping due
to diminishing curriculum coverage, while additional phases could still
improve the policy. The original WebRL phase-8 result uses a larger
interaction budget and is not evidence that the baseline converged.

\paragraph{Phase $K$: Environment Interaction and Data Collection.}
At the start of phase $K$, the current model $\text{RL-agent } K$ is deployed on the task distribution $\text{Task Set for Phase } K$ within the environment $\text{Env}$. This stage gathers diverse trajectories under the current policy's capabilities.

\begin{enumerate}
    \item \textbf{Rollout Execution.} The agent performs interactive rollouts within $\text{Env}$, producing sequences of states, actions, and observations that reflect both successful behaviors and failure cases.
    
\item \textbf{Trace Processing and Buffer Integration.} Newly generated rollouts are passed to the \emph{Trace Processing} module for annotation and filtering.
\begin{itemize}
    \item \textit{Subgoal Annotation:} We utilize LLM-based \textbf{subgoal checkers}\footnote{We adopt few-shot based Gemini-2.5-pro and engineered prompts} to verify the completion status of intermediate objectives at every step. These binary checks are interpolated to generate dense \textbf{subgoal progress scores} ($p_t^\ast$), which serve as regression targets for the Potential Critic.
    \item \textit{Perplexity Filtering:} To ensure the actor learns from high-quality data, we evaluate the \textbf{perplexity} of the current actor policy on the collected traces. Trajectories exhibiting perplexity above a dynamic threshold are discarded as out-of-distribution or irrational behavior.
    \item \textit{Training Target Assignment:} The filtered data is processed to support the hierarchical optimization in Algorithm~\ref{alg:mira-rl}. First, the \textbf{Potential Critic} is trained on the progress labels $p_t^\ast$. Second, the \textbf{Main Value Critic} is updated using the sparse terminal outcomes (ORM labels). Finally, using the updated potential landscape, we compute shaped rewards and the resulting \textbf{Monte-Carlo (MC) returns} ($G_t$) to optimize the \textbf{Actor}. For each phase, we perform one epoch of RL training over the collected data.
\end{itemize}
This produces an updated and temporally coherent \textbf{Training Data} corpus that captures the agent's evolving experience distribution while filtering out noise.
\end{enumerate}

\paragraph{Offline RL Training: Model Patching via Batched Updates.}
The aggregated $\text{Training Data}$ is then used to refine the model parameters through \emph{offline batched RL}, effectively ``patching'' the agent between deployment rounds. The training phase minimizes the regression-style policy loss $\mathcal{L}_\pi(\theta)$ from Eq.~\eqref{eq:mse_objective}, along with critic updates from Eqs.~\eqref{eq:potential_regression} and \eqref{eq:critic_value}. The result is a newly optimized policy $\text{RL-agent } K\!+\!1$, which inherits prior knowledge while incorporating corrections from the latest rollout data. Offline RL offers two practical benefits: (1) it stabilizes learning by decoupling environment noise from gradient updates, and (2) it enables fine-grained policy improvements at scale by batching large numbers of stored trajectories, including previously failed ones, into a consistent optimization phase.

\paragraph{Online Curriculum Generation: Failure-Driven Task Evolution.}
Once the offline phase concludes, the system performs targeted task selection to maintain progressive learning difficulty:

\begin{enumerate}
    \item \textbf{Task Generation ($\text{Task Gen.}$).} We utilize a pre-generated pool of 1{,}573 feasible tasks, authored by an LLM generator (Gemini-2.5-pro) seeded with domain descriptions and filtered for feasibility before use; generated tasks are further screened against the WebArena-Lite evaluation set to eliminate overlap. The $\text{Task Gen.}$ module analyzes the $\text{Failure}$ traces from Phase $K$ and resamples tasks from this pool that exhibit high similarity to the failed instances. In practice, we use larger models such as Gemini-2.5-flash to identify semantic similarity, limit the top-$K$ similar tasks per failure trace, and permit training-task reuse across phases. For rare or unbalanced task categories where the pool coverage is sparse (e.g., ``Map'' navigation), we synthesize new instructions by perturbing failure samples to ensure balanced distribution.
    
    \item \textbf{Adaptive Phase Transition.} The resulting $\text{Task Set for Phase } K\!+\!1$ becomes the curriculum for the next phase. The refined agent $\text{RL-agent } K\!+\!1$ is then redeployed, closing the loop between environment feedback and policy improvement.
\end{enumerate}

\paragraph{Iterative Curriculum Refinement.}
This iterative process forms a continual adaptation loop:
\[
\text{Phase } K \;\xrightarrow[\text{Failures}]{\text{Task Gen.}}\; \text{Phase } K\!+\!1,
\]
where each cycle explicitly leverages failure cases as learning signals. Over successive phases, the curriculum automatically evolves from simple to complex tasks, while the policy progressively internalizes successful behaviors through potential-shaped, subgoal-oriented offline RL. This cyclical design ensures that MiRA-RL maintains long-horizon learning efficiency while continually repairing and extending the model's competence boundary.

\begin{algorithm}[t]
\caption{MiRA Outer Loop: Curriculum Generation \& Experience Replay}
\label{alg:mira_curriculum}
\begin{algorithmic}[1]

\State \textbf{Input:} Initial set $I_0$, task pool $I_{\text{pool}}$, SFT model $\mathcal{M}_{\text{SFT}}$, Envs, Reward Model $\mathcal{M}_{\text{ORM}}$
\State \textbf{Initialize:} $\pi_1 \gets \mathcal{M}_{\text{SFT}}$, Critics $V_\phi, P_\psi$, Buffer $\mathcal{B} \gets \emptyset$, Failure Set $\mathcal{F} \gets \emptyset$, $I_{\text{train}} \gets I_0$

\For{Phase $k = 1$ to $N$}
    \State Collect trajectories $D_{\text{roll}}$ via $\pi_k$ on $I_{\text{train}}$; label successes with $\mathcal{M}_{\text{ORM}}$
    \State Split $D_{\text{roll}} \to D_{\text{success}} \cup D_{\text{fail}}$
    \State Update $\mathcal{B} \gets \mathcal{B} \cup D_{\text{success}}$ and $\mathcal{F} \gets \mathcal{F} \cup \{\text{instructions of } D_{\text{fail}}\}$

    \State Calculate $\text{ppl}(\tau)$ for $\tau \in \mathcal{B}$ using current $\pi_k$
    \State Filter $D_{\text{exp}} \gets \big\{\tau \in \mathcal{B} \mid 1/0.9 \le \text{ppl}(\tau) \le 1/0.5 \big\}$
    \State Combine data: $D_{\text{train}} \gets D_{\text{roll}} \cup D_{\text{exp}}$

    \State $(\pi_{k+1}, V_\phi, P_\psi) \gets \textsc{MiRA-RL}(D_{\text{train}}, V_\phi, P_\psi, \pi_k, \dots)$

    \State $I_{\text{sim}} \gets \text{Resample}(I_{\text{pool}})$ based on semantic similarity to $\mathcal{F}$
    \State $I_{\text{syn}} \gets \text{Generate}(\mathcal{F})$ for rare/unbalanced failure types
    \State $I_{\text{train}} \gets \text{Select}(I_{\text{sim}} \cup I_{\text{syn}})$
\EndFor

\end{algorithmic}
\end{algorithm}

\subsection{MiRA-RL Inner-Loop Algorithm}
\label{sec:apdx_algo}

Algorithm~\ref{alg:mira-rl} summarizes the inner-loop actor optimization with potential-based reward shaping. Three phases alternate per iteration: critic updates, reward shaping \& return computation, and the MSE-regression actor update.

\begin{algorithm}[h]
\caption{\textsc{MiRA-RL}: Actor Optimization with Potential-Based Reward Shaping (Inner Loop)}
\label{alg:mira-rl}
\begin{algorithmic}[1]
\Require Training batch $\mathcal{D}_{\text{train}}$; value critic $V_\phi$; potential critic $P_\psi$; current policy $\pi_\theta$; reference policy $\pi_{\mathrm{ref}}$; discount $\gamma$; temperature $\beta$; mixing $\lambda$; learning rates $\eta$.
\vspace{0.25em}

\State \textbf{Initialize} local optimization step counter $t \gets 0$
\While{optimization not converged}
  \State \textbf{Phase A: Critic updates}
  \For{$i=1$ to $N_c$}
    \State Sample minibatch $\mathcal{B}_c \subset \mathcal{D}_{\text{train}}$
    \State Update Value Critic $\mathcal{L}_V(\phi)$ via Eq.~\eqref{eq:critic_value}
    \State Update Potential Critic $\mathcal{L}_P(\psi)$ via Eq.~\eqref{eq:potential_regression}
    \State $\phi \gets \phi - \eta_V \nabla \mathcal{L}_V$, \quad $\psi \gets \psi - \eta_P \nabla \mathcal{L}_P$
  \EndFor

  \State \textbf{Phase B: Reward Shaping \& Returns}
  \State Sample minibatch $\mathcal{B}_a \subset \mathcal{D}_{\text{train}}$
  \For{each trajectory $\tau \in \mathcal{B}_a$}
    \State Calculate potential diffs: $r'_t \gets r_t + \alpha(P_\psi(s_{t+1}, g) - P_\psi(s_t, g))$
    \State Compute returns $G_t$ using shaped rewards $r'_t$
  \EndFor

  \State \textbf{Phase C: Actor Update}
  \For{each transition in $\mathcal{B}_a$}
    \State Compute Advantage $A_t^{\text{shaped}}$ (Eq.~\ref{eq:advantage}) mixing Monte-Carlo and Critic
    \State Compute log-ratio $\rho_t$ against $\pi_{\mathrm{ref}}$
  \EndFor
  \State Update $\theta \gets \theta - \eta_\pi \nabla_\theta \mathcal{L}_\pi(\theta)$ (minimizing Eq.~\ref{eq:mse_objective})
\EndWhile
\State \Return Updated parameters $(\pi_\theta, V_\phi, P_\psi)$

\end{algorithmic}
\end{algorithm}

\subsection{Experimental Details}
\label{sec:apdx_exp}
In this section, we provide a detailed breakdown of the experimental setup, including the compute infrastructure, the specific prompts used for subgoal decomposition, the hyperparameters for all baseline comparisons, and the system instructions that govern agent behavior and evaluation. While our motivation study (Section~\ref{sec:motivation}) analyzes the full 812-task WebArena benchmark to characterize failure modes at scale, our primary evaluation uses the 165-task WebArena-Lite subset. SFT initialization relies on separate cold-start tasks, and RL training uses generated tasks screened for overlap with the evaluation set.

\paragraph{Compute Infrastructure.}
All experiments were conducted on a single node equipped with 8 high-memory NVIDIA A100 GPUs (80GB each). A complete multi-phase training run of MiRA on Gemma3-12B---covering the potential critic warmup, the six-phase curriculum, and per-phase WebArena-Lite evaluation---takes approximately three days of wall-clock time on this setup (${\sim}580$ GPU-hours per full run). Per-phase evaluation over the 165 WebArena-Lite tasks completes in roughly 40 minutes. Including the five main-result seeds, the four ablation variants (w/o PC, w.\ KL, w/o Doubly Adv., AWR), and hyperparameter sensitivity sweeps ($\alpha$ grid, perplexity filtering), the full research project consumed approximately 6{,}000 GPU-hours.

\paragraph{Additional evaluation results and WebRL reproducibility.}
On the full 812-task WebArena benchmark, Gemma3-12B SFT, WebRL, and MiRA
achieve 18.8\%, 24.0\%, and 30.4\% Pass@1 SR, respectively. The full
suite contains difficult, unstable, and potentially infeasible cases; it
is not a fully independent environment. On WebArena-Lite, Llama3.1-8B
SFT, DigiRL, WebRL, and MiRA obtain 20.6$\pm$1.1, 30.3$\pm$1.4,
38.8$\pm$1.1, and 41.2$\pm$1.2\% Pass@1 SR (run-level SE).
Our 38.8\% WebRL score evaluates the released checkpoint in a unified
harness, while the original work reports 42.4\% in its configuration.
Repeated frozen-checkpoint evaluations in our harness give 37.6\%,
38.8\%, and 40.0\%; repeated ORM labeling changes the aggregate SR by
roughly 1--2 points. The matched Gemma3 comparison uses the same actor,
SFT initialization, reset/evaluation setup, and six-phase training budget
for WebRL and MiRA.

\paragraph{Inference cost.}
The subgoal generator, subgoal checker, and both critics are training-only:
at deployment, the actor uses the same instruction/history/HTML action
interface as matched baselines. Median episode-level input/output tokens
and action counts (with interquartile ranges) are:
\begin{table}[htbp]
\centering\footnotesize
\caption{Inference usage per episode: median [P25, P75].}
\label{tab:inference_usage}
\begin{tabular}{lccc}
\toprule
Policy & Input tokens & Output tokens & Actions \\
\midrule
Gemma3-SFT & 74k [42k, 118k] & 0.82k [0.51k, 1.31k] & 10 [6, 16] \\
Gemma3-WebRL & 86k [49k, 134k] & 0.95k [0.60k, 1.45k] & 12 [7, 18] \\
Gemma3-MiRA & 82k [47k, 128k] & 0.91k [0.58k, 1.38k] & 11 [7, 17] \\
\bottomrule
\end{tabular}
\end{table}

\paragraph{Subgoal Generation Examples and Decomposition.}
\label{sec:apdx_subgoaleg}
To bridge the gap between high-level user intents and low-level browser actions, MiRA employs a multimodal decomposition strategy. We utilize \textbf{Gemini-2.5-pro} as the teacher model due to its strong visual reasoning capabilities. As shown in Figure~\ref{fig:subgoal-prompt}, the model is prompted with the task instruction and the initial webpage screenshot, and is strictly instructed to decompose the task into a fixed number of milestones (e.g., exactly 4 steps). This standardization ensures consistent temporal granularity across diverse tasks. Table~\ref{tab:subgoal_examples} presents qualitative examples of these generated subgoals across various WebArena environments (Reddit, GitLab, CMS, etc.), demonstrating the model's ability to identify logical checkpoints such as ``Maps to subreddit,'' ``Search for user,'' and ``Submit post.''

\renewcommand{\arraystretch}{1.1}

\newcommand{\tablist}[1]{%
    \begin{enumerate}[leftmargin=*, nosep, before=\vspace{2pt}, after=\vspace{2pt}]
        #1
    \end{enumerate}%
}

\begin{longtable}{p{0.12\textwidth} p{0.30\textwidth} p{0.52\textwidth}}
\captionsetup{justification=centering}
\caption{Examples of generated subgoals across WebArena domains.}
\label{tab:subgoal_examples} \\
\toprule
\footnotesize \textbf{Website} & \footnotesize \textbf{User Intent} & \footnotesize \textbf{Generated Subgoals} \\
\midrule
\endfirsthead

\multicolumn{3}{c}%
{{\bfseries \footnotesize \tablename\ \thetable{} -- continued from previous page}} \\
\toprule
\footnotesize \textbf{Website} & \footnotesize \textbf{User Intent} & \footnotesize \textbf{Generated Subgoals} \\
\midrule
\endhead

\midrule
\multicolumn{3}{r}{{\footnotesize Continued on next page}} \\
\bottomrule
\endfoot

\bottomrule
\endlastfoot

\footnotesize

\multirow{8}{*}{\textbf{Reddit}} 
& Ask for product recommendations for running shoes within a budget of \$100 in r/sports 
& \tablist{
    \item Navigate to the ``r/sports'' subreddit.
    \item Initiate the creation of a new post.
    \item Write a title and body asking for shoes under \$100.
    \item Submit the post.
} \\ 
\cmidrule{2-3}
& Like all submissions created by CameronKelsey in subreddit earthporn 
& \tablist{
    \item Navigate to the subreddit `earthporn'.
    \item Search for posts by user `CameronKelsey'.
    \item Click the `like` button on the first post.
    \item Repeat for all other visible posts.
} \\
\midrule

\multirow{9}{*}{\textbf{Gitlab}} 
& Create a new public project ``awesome-llms'' and add primer, convexegg, abishek as members 
& \tablist{
    \item Navigate to new project creation page.
    \item Enter ``awesome-llms'', set visibility to public.
    \item Finalize project creation.
    \item Add primer, convexegg, and abishek in members page.
} \\
\cmidrule{2-3}
& Display issues in \texttt{kkroening and ffmpeg-python} with labels related to questions 
& \tablist{
    \item Navigate to \texttt{kkroening/ffmpeg-python} repo.
    \item Go to the ``Issues'' tab.
    \item Open ``Labels'' filter menu.
    \item Select label related to ``questions''.
} \\
\midrule

\multirow{8}{*}{\makecell[tl]{\textbf{Shopping}\\\textbf{Admin}\\\textbf{(CMS)}}} 
& Tell me the grand total of invoice 000000002 
& \tablist{
    \item Navigate to sales/invoices section.
    \item Search for invoice 000000002.
    \item Open details for the invoice.
    \item Identify grand total amount.
} \\
\cmidrule{2-3}
& Tell me the status of my latest order and when will it arrive 
& \tablist{
    \item Navigate to ``My Account'' or ``Profile''.
    \item Go to ``Orders'' or ``Order History''.
    \item Identify most recent order.
    \item Extract status and estimated arrival date.
} \\
\midrule

\multirow{9}{*}{\textbf{Map}} 
& Min travel time by car from Animal Rescue League of Pittsburgh to Schenley park? 
& \tablist{
    \item Navigate to map website, access directions.
    \item Enter ``Animal Rescue League'' as start.
    \item Enter ``Schenley park'' as destination.
    \item Select car mode, find min travel time route.
} \\
\cmidrule{2-3}
& Tell me the coordinates of Apple Store near Pitt in DD format 
& \tablist{
    \item Search for ``Pitt'' on map website.
    \item Search for ``Apple Store'' nearby.
    \item Select the correct Apple Store.
    \item Obtain coordinates in DD format.
} \\
\midrule

\multirow{9}{*}{\makecell[tl]{\textbf{Shopping}\\\textbf{(OSS)}}} 
& Look up most recent models of XBox controllers released 2020-2021? 
& \tablist{
    \item Search `Xbox controller' on Wikipedia.
    \item Navigate to main page.
    \item Locate `Models' or `History' section.
    \item Identify controllers released 2020-2021.
} \\
\cmidrule{2-3}
& Show least expensive switch card holder with capacity of 15+ cards. 
& \tablist{
    \item Search for ``switch card holder''.
    \item Filter for capacity of 15 or more.
    \item Sort results by price (low to high).
    \item Select first item.
} \\

\end{longtable}

\paragraph{Hyperparameter Configuration}
To ensure a fair comparison, we align our hyperparameters with established baselines wherever possible. Table~\ref{tab:hyperparams} lists the training configurations for SFT, AWR, DigiRL, WebRL, and MiRA. For the actor and critic networks, we maintain a consistent learning rate of $1 \times 10^{-6}$ and a batch size of 128 across all RL methods. A key distinction for MiRA is the \textbf{Potential Critic}, which uses a higher learning rate of $2 \times 10^{-5}$ and is trained for 3 epochs. This adjustment is necessary because the Potential Critic performs a regression task (fitting dense progress scores) rather than standard value estimation, requiring stronger gradient updates to capture the subtle shape of the progress landscape effectively.

\begin{table}[t!]
\centering
\footnotesize
\captionsetup{justification=centering}
\caption{Hyperparameters used in baseline methods.}
\label{tab:hyperparams}
\begin{tabular}{l l c}
\toprule
\textbf{Method} & \textbf{Hyperparameter} & \textbf{Value} \\
\midrule
\multirow{6}{*}{SFT} 
& learning rate & $1\times10^{-5}$ \\
& lr scheduler type & cosine \\
& warmup ratio & 0.1 \\
& batch size & 128 \\
& training epoch & 1 \\
& cutoff length & 16384 \\
\midrule
\multirow{8}{*}{AWR}
& actor learning rate & $1\times10^{-6}$ \\
& actor lr scheduler type & constant \\
& critic learning rate & $1\times10^{-6}$ \\
& critic lr scheduler type & constant \\
& batch size & 128 \\
& discount factor & 0.9 \\
& actor training epoch & 1 \\
& critic training epoch & 1 \\
\midrule
\multirow{12}{*}{DigiRL}
& actor learning rate & $1\times10^{-6}$ \\
& actor lr scheduler type & constant \\
& critic learning rate & $1\times10^{-6}$ \\
& critic lr scheduler type & constant \\
& instruction value function lr & $1\times10^{-6}$ \\
& instr.\ value function scheduler & constant \\
& batch size & 128 \\
& discount factor & 0.9 \\
& actor training epoch & 1 \\
& critic training epoch & 1 \\
& instruction value function epoch & 1 \\
& rollout temperature & 1 \\
& replay buffer size & 100{,}000 \\
\midrule
\multirow{8}{*}{WebRL}
& actor learning rate & $1\times10^{-6}$ \\
& actor lr scheduler type & constant \\
& critic learning rate & $1\times10^{-6}$ \\
& critic lr scheduler type & constant \\
& batch size & 128 \\
& discount factor & 0.9 \\
& actor training epoch & 1 \\
& critic training epoch & 1 \\
& rollout temperature & 1 \\
\midrule
\multirow{12}{*}{MiRA}
& actor learning rate & $1\times10^{-6}$ \\
& actor lr scheduler type & constant \\
& critic learning rate & $1\times10^{-6}$ \\
& critic lr scheduler type & constant \\
& potential critic learning rate & $2\times10^{-5}$ \\
& potential critic lr scheduler type & constant \\
& potential critic training epoch & 3 \\
& batch size & 128 \\
& discount factor & 0.9 \\
& actor training epoch & 1 \\
& critic training epoch & 1 \\
& rollout temperature & 1 \\
\bottomrule
\end{tabular}
\end{table}

\begin{figure}[t!]
    \centering
    \includegraphics[width=1.0\linewidth]{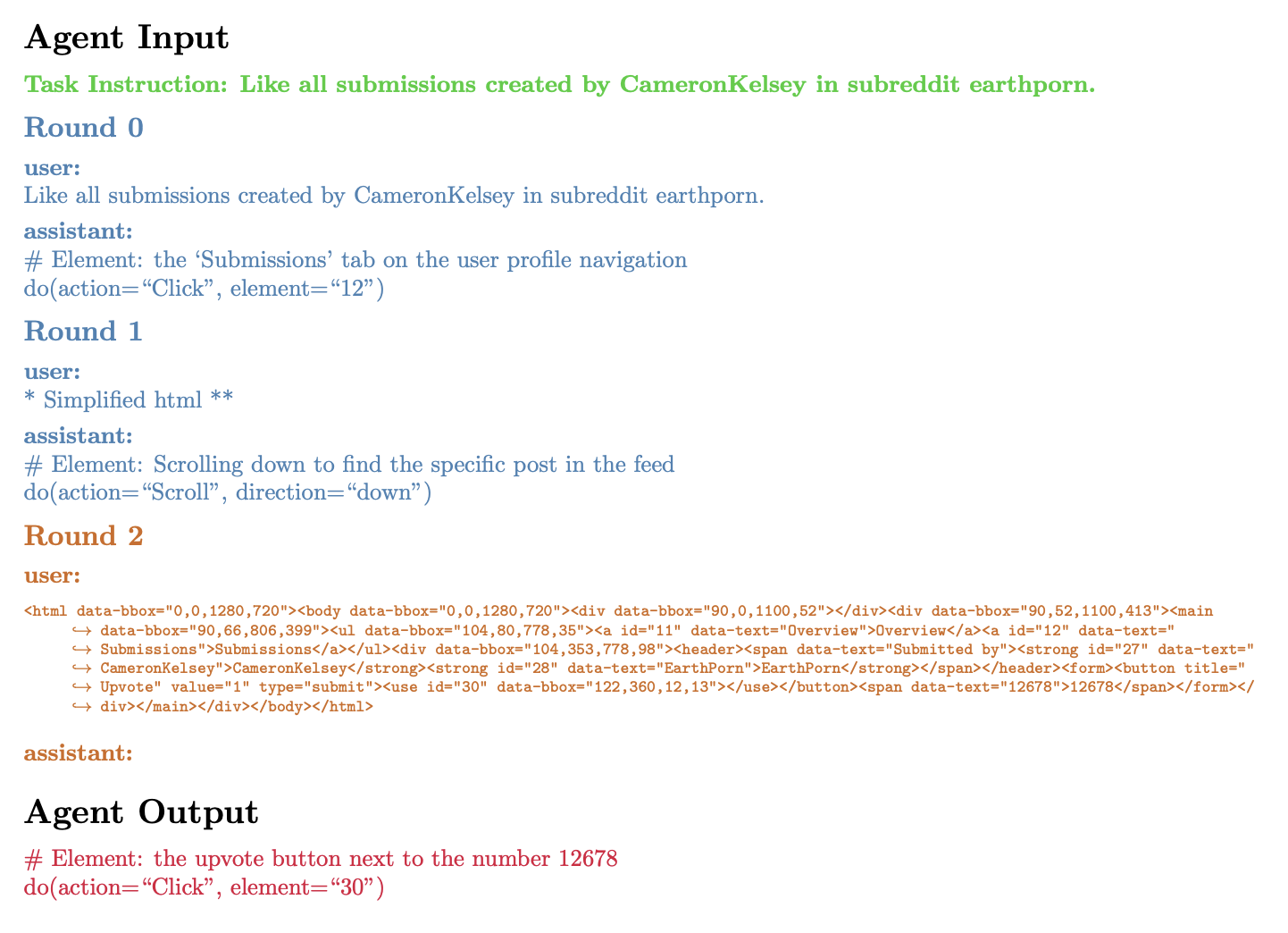}
    \caption{The input and output format of MiRA and baselines, where the input is composed of task instruction (in green), action history (in blue), and HTML of the current webpage (in orange). The output (in red) is the action taken on the current webpage.}
    \label{fig:apdx_agent}
\end{figure}

\paragraph{System Prompts for Execution and Evaluation}
The performance of web agents is heavily influenced by the precise formatting of their inputs and allowed outputs. Figure~\ref{fig:apdx_agent} visualizes the data flow, where the agent receives the task instruction, action history, and current HTML. To constrain the agent's output to valid executable actions, we employ the system prompt detailed in Figure~\ref{fig:system-prompt} for small open-weight models (e.g., Llama-3, Gemma). This prompt restricts responses to a predefined Python-style DSL (e.g., \texttt{do(...)}, \texttt{click(...)}) and explicitly forbids the hallucination of non-visible elements.

In contrast, when evaluating large proprietary models (e.g., GPT-4o, Gemini-2.5-Pro), we adopt the standard text-based accessibility tree prompt shown in Figure~\ref{fig:proprietary-prompt}. This configuration aligns with the default WebArena benchmark settings and prior works~\citep{chae2025web,qi2024webrl,wei2025webagent}, utilizing the stronger instruction-following capabilities of frontier models to parse structured element lists rather than simplified HTML. Finally, for reward modeling, we utilize the Outcome Reward Model (ORM) prompt shown in Figure~\ref{fig:orm-prompt}. This prompt instructs the judge to evaluate success based on the alignment between the user intent and the final page state, serving as the ground-truth signal for filtering successful trajectories during the warmup phase.

\begin{figure}[t!]
\centering
\footnotesize
\begin{tcolorbox}[colback=gray!3,colframe=black!40,title=System Prompt Used in small model baselines incl. MiRA]
\textbf{\# Setup} \\
You are a professional web browsing agent assistant that can fulfill user's high-level
instructions. Given simplified HTML of the browsed webpage at each step, you plan operations
in Python-style pseudo code using provided functions, or customize functions if necessary. \\[3pt]

\textbf{\# Predefined Functions}
\begin{verbatim}
def do(action, argument, element):
    """A single browsing operation on the webpage."""
def exit(message):
    """End the browsing process."""
def go_backward():
    """Go back to the previous page."""
def go_forward():
    """Go forward to the next page."""
\end{verbatim}

\textbf{Allowed actions:}
Click, Right Click, Type, Search, Hover, Scroll Up, Scroll Down, Press Enter,  
Switch Tab, Select Dropdown Option, Wait. \\[3pt]

\textbf{Examples:}
\begin{verbatim}
# Element: the `REPORTS' section on the left sidebar
do(action="Click", element="7")

# Element: the `Period' dropdown
do(action="Select Dropdown Option", argument="Month", element="20")
\end{verbatim}

\textbf{Rules:}
\begin{itemize}[nosep,leftmargin=10pt]
\item Only \textbf{one line of code} at a time.
\item Do not invent non-visible elements.
\item Use ``\# Element'' comments to indicate the chosen element.
\item Avoid loops; use if-else only when necessary.
\item If the page is loading, prefer ``Wait''.
\item Never use the browser address bar to navigate.
\end{itemize}

\textbf{User Prompt Template:}
\begin{verbatim}
Task Instruction: {instruction}
Action History: {action history}
The Current HTML: {html of last state}
\end{verbatim}
\end{tcolorbox}
\captionsetup{justification=centering}
\caption{The simplified system prompt employed in baseline agents.}
\label{fig:system-prompt}
\end{figure}

\begin{figure}[t!]
\centering
\footnotesize
\begin{tcolorbox}[colback=gray!3,colframe=black!40,title=System Prompt for Outcome Reward Model (ORM)]
\textbf{\# Role} \\
You are an expert in evaluating the performance of a website navigation agent. The agent is designed to help a human user navigate the website to complete a task. Given the user's intent, the agent's action history, and the final state of the screen, your goal is to decide whether the agent's execution is successful or not. You must respond with YES or NO. \\[3pt]

\textbf{\# User Prompt Template}
\begin{verbatim}
The User Intent:
{task instruction}

Action History:
{history of agent actions}

The Current Screenshot or Page:
{simplified html/screenshots of final state}
\end{verbatim}
\end{tcolorbox}
\caption{The system prompt used for the Outcome Reward Model (ORM) to verify trajectory success.}
\label{fig:orm-prompt}
\end{figure}

\begin{figure}[t!]
\centering
\footnotesize
\begin{tcolorbox}[colback=gray!3,colframe=black!40,title=System Prompt for Proprietary Models (GPT-4 / Gemini)]
\textbf{\# Role} \\
You are an autonomous intelligent agent tasked with navigating a web browser. You will be given web-based tasks and must accomplish them by issuing specific actions. \\[3pt]

\textbf{\# Input Information}
\begin{itemize}[nosep,leftmargin=10pt]
\item \textbf{Objective:} The user's task description.
\item \textbf{Observation:} A simplified accessibility tree representation of the webpage.
\item \textbf{Current URL:} The page you are currently navigating.
\item \textbf{Previous Action:} The action you just performed.
\end{itemize}

\textbf{\# Available Actions}
\begin{verbatim}
click [id]                  # Click on element with specific ID
type [id] [content] [0|1]   # Type content (0=no enter, 1=enter)
hover [id]                  # Hover over element
press [key_comb]            # Simulate key combination (e.g., "Ctrl+v")
scroll [up|down]            # Scroll the page
new_tab, close_tab          # Manage tabs
tab_focus [index]           # Switch to tab by index
goto [url]                  # Navigate to URL
go_back, go_forward         # History navigation
stop [answer]               # Complete task (with optional answer)
\end{verbatim}

\textbf{\# Response Format}
Your response must always start with the phrase ``In summary, the next action I will perform is'' followed by the action inside triple backticks.

\textbf{\# Examples}
\begin{verbatim}
OBSERVATION:
[1744] link `HP Inkjet Fax Machine'
[1757] button `Add to Cart'
URL: http://onestopmarket.com
OBJECTIVE: What is the price of the Fax Machine?

RESPONSE:
In summary, the next action I will perform is ```stop [$279.49]'''
\end{verbatim}

\begin{verbatim}
OBSERVATION:
[164] textbox `Search'
[171] button `Go'
URL: http://openstreetmap.org
OBJECTIVE: Show me hotels in Pittsburgh

RESPONSE:
In summary, the next action I will perform is
```type [164] [hotels in Pittsburgh] [1]'''
\end{verbatim}
\end{tcolorbox}
\caption{The standard system prompt (Set-of-Marks Style) used for large proprietary models (e.g., GPT-4, Gemini) in WebArena, utilizing a text-based accessibility tree observation space.}
\label{fig:proprietary-prompt}
\end{figure}

\begin{figure}[t!]
\centering
\footnotesize
\begin{tcolorbox}[colback=gray!3,colframe=black!40,title=Prompt for Subgoal Generation]
\textbf{\# Role} \\
You are an assistant inside the WebArena benchmark environment. WebArena simulates websites such as GitLab, Wikipedia, shopping portals, and admin dashboards. Tasks are solved only through browser interactions (clicks, typing, navigation)—never by recalling outside world knowledge. Your job is to decompose each task intent into EXACTLY 4 concrete, high-level subtasks. Always provide exactly 4 numbered steps, no more and no less. \\[3pt]

\textbf{\# Example 1: GitLab}
\begin{verbatim}
Task: Create a private JEKYLL repository called 'awesome_project'.
Screenshots: [Image1], [Image2], [Image3]

Answer:
1. Access GitLab and locate 'Create New'
2. Select the JEKYLL template from the template menu
3. Enter repo name 'awesome_project' and set it private
4. Create repository and confirm landing on its homepage
\end{verbatim}

\textbf{\# Example 2: Map/Wiki}
\begin{verbatim}
Task: Find the page of the college(s) where *The Chair* was filmed in
Pennsylvania other than those in Pittsburgh on the map.
Screenshots: [Image1], [Image2], [Image3]

Answer:
1. Identify correct Wikipedia page for *The Chair* (2021 TV series) and
   locate filming colleges in Pennsylvania (excluding Pittsburgh)
2. Copy the college name from the Wikipedia page
3. Open the map website and search the copied name
4. Navigate to the college's location on the map
\end{verbatim}

\textbf{\# User Prompt Template}
\begin{verbatim}
=====Your Turn=====
Q: TASK: {task_intent}
Screenshot: {task_initial_screenshot}
Decompose into exactly 4 milestones.
A: <SUBGOALS>
\end{verbatim}
\end{tcolorbox}
\caption{The multimodal prompt used to decompose tasks into four structured subgoals (Taking fixed 4 subgoals in total as example) via Gemini-2.5-pro.}
\label{fig:subgoal-prompt}
\end{figure}

\subsection{MiRA Agent Design and Rationale}
\label{sec:apdx_agent_design}

\subsubsection{Observation and Action Space}
The MiRA agent operates within a structured environment designed to bridge the gap between high-level user instructions and low-level DOM interactions. The agent's observation space consists of three key components:
\begin{itemize}
    \item \textbf{User Instruction:} The natural language goal provided by the user.
    \item \textbf{Action History:} A chronological record of past actions, providing essential context for tracking subgoal progress.
    \item \textbf{Webpage HTML:} A simplified representation of the current webpage's HTML. To facilitate precise manipulation, we prune the DOM structure and assign unique element IDs to all interactive candidates.
\end{itemize}

The action space is designed to be granular yet comprehensive, enabling the execution of complex subgoals. The defined actions include:
\begin{itemize}[noitemsep]
    \item \textbf{Interaction:} \textit{Click}, \textit{Hover}, \textit{Type}, \textit{Search} (type and enter), \textit{Select dropdown option}.
    \item \textbf{Navigation:} \textit{Goto}, \textit{Go back}, \textit{Go forward}, \textit{Scroll}.
    \item \textbf{Tab Management:} \textit{New tab}, \textit{Tab focus}, \textit{Close tab}.
    \item \textbf{System:} \textit{Press} (key combinations), \textit{Exit} (terminate and respond).
\end{itemize}

To enhance interpretability and grounding, the agent appends metadata to its actions: a \texttt{\# Element:} tag describing the target DOM element, and a \texttt{\# Note:} tag citing supporting information from the webpage content.

\subsubsection{Training Framework and Rationale}
The design of the MiRA training pipeline addresses the challenge of learning robust policies from sparse rewards. The process follows a self-evolving curriculum:

\paragraph{Initialization via Supervised Fine-Tuning (SFT):}
We establish a baseline web-navigation capability by initializing the actor
from an SFT checkpoint trained on a separate set of cold-start tasks,
strictly disjoint from the WebArena-Lite evaluation tasks.
This checkpoint also initializes the backbone of the critics
(with newly initialized value heads), and the policy populates
the initial training replay and failure sets.
For each backbone, all matched RL methods use the same SFT initialization.


\paragraph{Self-Evolving Curriculum Reinforcement Learning:}
To progressively enhance performance, we employ a curriculum-based reinforcement learning approach. In each phase, we sample up to 400 new instructions from the LLM-generated task pool, with Gemini-2.5-flash filtering for complexity criteria consistent with the agent's current capability frontier. To prevent train-evaluation overlap, we explicitly exclude any generated instruction whose semantic similarity to any WebArena-Lite evaluation task exceeds a pre-defined threshold, computed via sentence-level embedding similarity. The actor and critic are updated using a hybrid data mixture:
\begin{enumerate}
    \item Newly generated interaction trajectories from the current curriculum phase.
    \item Historical high-quality data from the replay buffer, specifically filtering for trajectories with perplexity between $1/0.9$ and $1/0.5$.
\end{enumerate}
This dual-source training strategy ensures the agent retains previously learned behaviors while adapting to increasingly complex subgoals.

\subsection{Extra Ablation Studies}
\label{sec:apdx_ablation}

\begin{table}[htbp]
\centering
\captionsetup{justification=centering}
\caption{Impact of replay buffer filtering based on perplexity (PPL) and rank score ($1/\mathrm{PPL}$).}
\label{tab:perplexity_impact}
\begin{tabular}{lcccc}
\toprule
\textbf{Rank Score ($1/\mathrm{PPL}$)} & $(0.9, 1.0]$ & $\mathbf{[0.5, 0.9]}$ & $[0.0, 0.5)$ & $[0.0, 1.0]$ \\
\midrule
\textbf{PPL Range} & $[1, \tfrac{1}{0.9})$ & $\mathbf{[\tfrac{1}{0.9}, \tfrac{1}{0.5}]}$ & $(\tfrac{1}{0.5}, \infty)$ & Full Range \\
\midrule
SR (\%)   & 27.9 & \textbf{36.4} & 23.6 & 29.1 \\
\bottomrule
\end{tabular}
\end{table}

\noindent \textbf{The impact of reward shaping ($\alpha$).}
We perform a grid search over $\alpha \in [0.0, 0.8]$ using a held-out validation set of WebArena-Lite tasks. As shown in Table~\ref{tab:alpha_abl}, very small shaping factors (0.0–0.1) behave similarly to the sparse-reward baseline, yielding only 30.9--31.5\% success. In this regime, the auxiliary signal is too weak to overcome reward sparsity and thus behaves largely as noise.

Performance peaks at $\alpha = 0.3$, achieving \textbf{36.4\%}, confirming that moderate shaping effectively accelerates credit assignment without drowning out the terminal reward. However, as $\alpha$ increases (0.5–0.8), performance drops sharply to 28.5–25.5\%. Large shaping factors distort the reward landscape, causing the agent to overfit to auxiliary subgoal progress rather than true task completion. These findings highlight the necessity of calibrated shaping magnitudes.

\begin{wrapfigure}{r}{0.45\textwidth}
    \centering
    \captionsetup{justification=centering}
    \captionsetup{font=footnotesize}
    \captionof{table}{Impact of reward shaping factor $\alpha$.}
    \footnotesize
    \begin{tabular}{lcccccc}
        \toprule
        $\alpha$ & 0.0 & 0.1 & \textbf{0.3} & 0.5 & 0.8 \\
        \midrule
        SR (\%) & 30.9 & 31.5 & \textbf{36.4} & 28.5 & 25.5 \\
        \bottomrule
    \end{tabular}
    \label{tab:alpha_abl}
\end{wrapfigure}

\noindent \textbf{The impact of perplexity filtering.}
Table~\ref{tab:perplexity_impact} evaluates how different perplexity bands affect data usefulness. Training exclusively on low-perplexity (highly predictable) samples yields only 27.9\% success, suggesting limited learning value from ``too easy'' trajectories. Conversely, using only the high-perplexity tail produces the worst performance (23.6\%), indicating that noisy or out-of-distribution states destabilize the value critic.

The best performance arises from filtering to the moderate-perplexity range $[1/0.9,1/0.5]$ (i.e., rank score $[0.5,0.9]$), achieving \textbf{36.4\%}. These ``borderline-difficult'' transitions provide the most informative gradients—neither trivial nor erratic—leading to more robust and sample-efficient learning.

\paragraph{Critic-target decoupling.}
The following controlled study compares outcome-only, entangled shaped-return,
and MiRA's decoupled outcome/potential critic targets. The AUROC measures
\emph{terminal-outcome prediction by the value critic}, not the AUROC of
graded subgoal completion reported in Section~\ref{sec:sg_gen}.
\begin{table}[htbp]
\centering\footnotesize
\caption{Outcome-critic target ablation (reported controlled-study SR).}
\label{tab:critic_target_ablation}
\begin{tabular}{lcc}
\toprule
Target design & Outcome-value AUROC & SR (\%) \\
\midrule
Outcome only, no progress & 0.78 & 30.9 \\
Entangled shaped-return critic & 0.70 & 32.8 \\
Decoupled MiRA & \textbf{0.81} & \textbf{36.4} \\
\bottomrule
\end{tabular}
\end{table}

\paragraph{Subgoal reliability versus executor and granularity.}
Holding the subgoal definition and checker fixed, stronger executors
more reliably reach intended milestones; varying the number of milestones
illustrates the precision--recall trade-off. These analyses evaluate
milestone--success agreement, not a head-to-head test of alternate
subgoal generators.
\begin{table}[htbp]
\centering\footnotesize
\caption{Executor comparison: same milestone definition and checker.}
\label{tab:milestone_executor}
\begin{tabular}{lcccc}
\toprule
Executor & Exact F1 & Precision & Recall & Graded AUROC \\
\midrule
Llama3.1-8B & 0.61 & 0.74 & 0.52 & 0.79 \\
Gemma3-12B & 0.68 & 0.79 & 0.60 & 0.84 \\
Gemini-2.5-Pro & 0.82 & 0.90 & 0.76 & 0.91 \\
\bottomrule
\end{tabular}
\end{table}
\begin{table}[htbp]
\centering\footnotesize
\caption{Number of milestones on Gemma3-12B: calibration trade-off.}
\label{tab:milestone_granularity}
\begin{tabular}{ccccc}
\toprule
Milestones & Exact F1 & Precision & Recall & Graded AUROC \\
\midrule
3 & 0.66 & 0.73 & 0.61 & 0.80 \\
4 & 0.68 & 0.79 & 0.60 & 0.84 \\
5 & 0.69 & 0.82 & 0.60 & 0.86 \\
6 & 0.67 & 0.85 & 0.56 & 0.87 \\
\bottomrule
\end{tabular}
\end{table}

\noindent\textbf{Interpretation of shaping sensitivity.}
The 36.4\% values in these controlled sensitivity studies are not the
six-phase main-result figure (43.0\%); the main-result and sensitivity
settings should not be conflated. The $\alpha$ sweep already reported
above shows a peak at $\alpha=0.3$ and performance degradation at
larger weights, as expected for an imperfect auxiliary objective.

\subsection{Potential Critic Training via Supervised Progress Regression}
\label{apdx:potential_training}

\begin{algorithm}[h]
\caption{Training the Potential Critic $P_\psi$}
\label{alg:potential_critic_final}
\begin{algorithmic}[1]
\Require Dataset of trajectories $\mathcal{D}$; goal set $\mathcal{G}=\{g_1,\dots,g_K,g_{\mathrm{final}}\}$
\Ensure Trained potential critic $P_\psi(s,g)$

\State $\mathcal{S} \gets \emptyset$ \Comment{Supervision set for regression}

\For{each trajectory $\tau\in\mathcal{D}$}
    \If{trajectory does \emph{not} reach the final goal} 
        \State \textbf{continue} \Comment{Use only positive traces to ensure correlation with success}
    \EndIf

    \State Extract subgoal-completion vector $\mathbf{z}_t$ for all $t$
    \State Identify key steps $t_1,\dots,t_K$ where cumulative completion increases

    \State Compute continuous progress labels $p_t^\ast$ by linear interpolation:
    \[
    p_t^\ast = (1-\alpha_t)\tfrac{j}{K} + \alpha_t\tfrac{j+1}{K},
    \quad \alpha_t=\frac{t - t_j}{t_{j+1}-t_j}
    \]
    with the final segment anchored to the true trajectory end $T$.

    \For{each timestep $t$}
        \State Construct progress-aware state $s_t = (h_{1:t-1}, o_t, g)$
        \State Add $(s_t, g, p_t^\ast)$ to $\mathcal{S}$
    \EndFor
\EndFor

\Statex
\State \textbf{/* Supervised learning */}
\State Train the potential critic via MSE regression:
\[
\mathcal{L}_P(\psi)=\frac{1}{|\mathcal{S}|}\sum_{(s_t,g,p_t^\ast)\in\mathcal{S}}
\left(P_\psi(s_t,g)-p_t^\ast\right)^2.
\]

\State Return trained $P_\psi$.
\end{algorithmic}
\end{algorithm}

A central challenge in long-horizon web tasks is that the environment provides only a sparse terminal reward—typically a single binary success or failure at the end of a multi-step trajectory. Such extreme sparsity makes it difficult for an RL agent to learn meaningful credit assignment, particularly when intermediate actions influence success in subtle, delayed ways. To mitigate this, MiRA introduces a \emph{potential critic} $P_\psi(s,g)$ that predicts a dense, continuous notion of task progress. Instead of relying on handcrafted shaping potentials, which are notoriously brittle in heterogeneous environments like WebArena, the potential critic is learned directly from data via supervised regression, yielding a shaping landscape that reflects the procedural structure of successful trajectories.

The construction of the regression targets is grounded in the SubGoal Checker, which emits a binary subgoal-completion vector $\mathbf{z}_t$ at each timestep. From this signal, we detect the key timesteps $t_1, t_2, \ldots, t_K$ at which cumulative progress increases. Between any pair of these key steps, we interpolate the progress value linearly so that the label $p_t^\ast$ increases smoothly from $j/K$ to $(j{+}1)/K$. This interpolation converts the discrete, staircase-like profile of subgoal completions into a continuous supervision signal defined over the entire trajectory. For the final segment, we intentionally anchor the interpolation interval to the true end of the trajectory rather than the moment the final subgoal is detected; this ensures that the learned potential continues to rise during late-stage ``cleanup'' actions such as verification, scrolling, and form submission, which contribute meaningfully to task resolution despite not triggering additional subgoal boundaries.

An important modeling decision is to fit the potential critic exclusively on \emph{positive} trajectories—those that ultimately reach the final goal. This ensures that the progress targets are always aligned with successful procedural structure. Using failed trajectories as supervision would dilute this relationship, as these episodes often terminate prematurely or contain exploratory detours that do not correspond to genuine forward progress. Restricting the regression dataset to successful episodes produces a potential function that implicitly captures the common semantic subsequences shared across successful attempts, creating a shaping landscape that rises precisely along paths known to lead to final success.

Once progress labels $p_t^\ast$ have been computed, the potential critic is trained to regress these values using a standard MSE objective:
\[
\mathcal{L}_P(\psi)
= \mathbb{E}_{(s_t,g,p_t^\ast)} \bigl[\|P_\psi(s_t,g) - p_t^\ast\|^2\bigr].
\]
The critic is implemented by augmenting a pretrained LLM with a regression head, enabling it to leverage both the semantic structure of the goal instruction and the action–observation history of the trajectory. Because the target labels are smooth and monotonic, optimization is stable, and the learned potential landscape provides a faithful surrogate of distance-to-go long before RL finetuning begins.

During reinforcement learning, the potential critic contributes an auxiliary shaping reward of the form
\[
r^{(\mathrm{aux})}_t
= \alpha \bigl[ P_\psi(s_{t+1},g) - P_\psi(s_t,g) \bigr],
\]
which is added to the environmental reward when computing advantage targets. This shaping term effectively supplies a local gradient that encourages the agent to continue moving ``forward'' in the learned progress space. The primary value critic $V_\phi$ remains trained on terminal outcomes, so it measures a distinct quantity from the potential critic. This separation anchors the actor surrogate to task success but does not ensure policy invariance: the auxiliary term can change the learned policy. Its usefulness and limitations are assessed in the shaping-weight and critic-target ablations in Appendix~\ref{sec:apdx_ablation}.

\noindent \textbf{Trajectory Selection and Quality Filtering.} To ensure the potential critic learns from robust and efficient behaviors, we apply strict quality controls to the gathered rollouts. First, we utilize an Outcome Reward Model (ORM) trained via WebRL~\citep{qi2024webrl} to verify the success of each trajectory collected from both the WebRL (Llama3-8b) and baseline RL agents (initialized from the same Gemma SFT checkpoint as MiRA). The task set remains identical to that used in the multi-phase running on the baseline methods. We then filter the verified successes to eliminate signs of suboptimal agency: we remove trajectories containing repetitive loops (defined as identical actions repeated more than 5 times) and exclude inefficient rollouts by retaining only those with a total length of less than 15 steps. This curation process yields a dataset of positive execution traces that represent more direct and logical paths to task completion, ensuring the learned potential landscape encourages efficiency rather than circuitous wandering. Then the potential critic continues to be trained jointly with the main critic and actor.

In summary, supervised progress regression allows MiRA to construct a dense potential landscape grounded in the structure of successful trajectories. This landscape provides smooth intermediate rewards that fill the gaps left by sparse feedback, enabling efficient learning of long-horizon procedural behaviors that would otherwise be inaccessible to standard reinforcement learning.

\subsection{Derivation of the Optimal Policy in KL-Regularized RL}
\label{sec:apdx_RL_adv}

We start with the objective function for KL-regularized Reinforcement Learning (commonly used in RLHF). The goal is to maximize the expected reward while minimizing the KL-divergence from a reference policy $\pi_{\text{ref}}$:

\begin{equation}
    \max_\pi \mathbb{E}_{\pi} \left[ r(s,a) - \beta \log \frac{\pi(a|s)}{\pi_{\text{ref}}(a|s)} \right]
\end{equation}

The closed-form solution for the optimal policy $\pi^*$ is given by the Boltzmann distribution:

\begin{equation}
    \pi^*(a|s) = \frac{1}{Z(s)} \pi_{\text{ref}}(a|s) \exp\left( \frac{Q^*(s,a)}{\beta} \right)
\end{equation}

where $Z(s)$ is the partition function (normalization constant). In the maximum entropy framework, the optimal value function $V^*(s)$ relates to the partition function as follows:

\begin{equation}
    V^*(s) = \beta \log Z(s) \implies Z(s) = \exp\left( \frac{V^*(s)}{\beta} \right)
\end{equation}

Recall the definition of the Advantage function: $A^*(s,a) = Q^*(s,a) - V^*(s)$. We can substitute $Q^*(s,a) = A^*(s,a) + V^*(s)$ back into the policy equation:

\begin{align*}
    \pi^*(a|s) &= \frac{1}{Z(s)} \pi_{\text{ref}}(a|s) \exp\left( \frac{A^*(s,a) + V^*(s)}{\beta} \right) \\
    &= \frac{1}{Z(s)} \pi_{\text{ref}}(a|s) \exp\left( \frac{A^*(s,a)}{\beta} \right) \exp\left( \frac{V^*(s)}{\beta} \right)
\end{align*}

Substituting $Z(s) = \exp\left( \frac{V^*(s)}{\beta} \right)$, the normalization terms cancel out:

\begin{align*}
    \pi^*(a|s) &= \frac{1}{\exp\left( \frac{V^*(s)}{\beta} \right)} \pi_{\text{ref}}(a|s) \exp\left( \frac{A^*(s,a)}{\beta} \right) \exp\left( \frac{V^*(s)}{\beta} \right) \\
    \pi^*(a|s) &= \pi_{\text{ref}}(a|s) \exp\left( \frac{1}{\beta} A^*(s,a) \right)
\end{align*}

\subsection{KL Divergence Formulation for Policy Update}
\label{sec:apdx_mse_vs_kl}

Many prior studies use KL divergence to measure the distance between two policies. When KL divergence is used to optimize $\pi_\theta$ toward $\pi^*$, the optimization goal becomes:
\begin{align}
& \arg\min_{\theta} \, \mathbb{E}_{s \sim d(s)} \left[ D_{\mathrm{KL}}\left( \pi^*(\cdot|s,I) \,\|\, \pi_\theta(\cdot|s,I) \right) \right] \nonumber \\[6pt]
&= \arg\max_{\theta} \, \mathbb{E}_{s \sim d(s)} \mathbb{E}_{a \sim \pi^*(a|s,I)} \left[ \log \pi_\theta(a|s,I) \right] \label{eq:kl_step1} \\[6pt]
&= \arg\max_{\theta} \, \mathbb{E}_{s \sim d(s)} \int_a \pi_{\mathrm{ref}}(a|s,I) \exp\!\left( \frac{1}{\beta} A^*(s,a,I) \right) \log \pi_\theta(a|s,I) \, \mathrm{d}a \label{eq:kl_step2} \\[6pt]
&= \arg\max_{\theta} \, \mathbb{E}_{s \sim d(s)} \mathbb{E}_{a \sim \pi_{\mathrm{ref}}(a|s,I)} \left[ \log \pi_\theta(a|s,I) \cdot \exp\!\left( \frac{1}{\beta} A^*(s,a,I) \right) \right], \label{eq:kl_final}
\end{align}
where $d(s)$ is a distribution over states, and we have substituted the closed-form optimal policy $\pi^*(a|s,I) = \pi_{\mathrm{ref}}(a|s,I) \exp\!\left( \frac{1}{\beta} A^*(s,a,I) \right)$ from Appendix~\ref{sec:apdx_RL_adv}.

Direct optimization of the forward KL objective requires expectations under the target policy or appropriate importance weighting. This can be challenging when learning from heterogeneous off-policy replay data. In contrast, our MSE log-ratio objective (Eq.~\ref{eq:mse_objective}) directly regresses signed surrogate targets on replay samples, allowing both positive and negative adjustments to sampled action log-probabilities. This provides a practical alternative for our off-policy training setting, although it does not guarantee exact policy improvement.

\subsection{Extended Discussion}
\label{sec:apdx_discussion}

In this work, we addressed the critical planning bottleneck in autonomous web agents by introducing MiRA, a unified framework that synergizes explicit milestone-based reasoning with potential-based reinforcement learning. By transforming high-level intent into verifiable subgoals, we demonstrated that open-source RL policies can overcome the horizon truncation failures pervasive in current benchmarks. Beyond the empirical gains on WebArena-Lite, our findings open several avenues for discussion regarding the nature of reward shaping, the reliability of intermediate supervision, and the future of self-evolving autonomy.

\paragraph{Connection to Process Reward Models (PRM).}
Conceptually, our training methodology can be viewed as a form of semi-supervised Process Reward Model (PRM) learning. Unlike traditional PRMs that rely on expensive human annotations for every step of a chain-of-thought, MiRA synthesizes its own ``process supervision'' by generating semantic subgoals via an advanced teacher model and grounding them in execution traces. The potential critic $P_\psi$ effectively acts as a learned value function over these synthesized milestones, densifying the sparse outcome signal. This aligns with recent findings in mathematical reasoning \citep{lightman2023let}, suggesting that verifying intermediate steps (subgoals) is often easier and more sample-efficient than verifying final answers alone. Our work extends this principle to the noisy, non-stationary domain of web navigation, proving that automated process supervision is a viable path to scaling RL for complex agents.

\paragraph{The ``Solid Grounding'' Trade-off.}
The efficacy of our potential-based shaping hinges entirely on the validity of the generated subgoals. As shown in our ablation studies, rewarding \textbf{solid} subgoals—those the agent can robustly perceive and achieve—creates a reliable curriculum, forming a regression curve of progress (e.g., in Gitlab tasks) that pulls the agent deeper into the episode. However, this reliance introduces a subtle \emph{amplification bias}. Because the auxiliary reward is triggered only upon successful grounding of a milestone, the system heavily favors trajectories that master the early stages of a task. If an agent fails to ground the initial subgoals due to extreme exploration difficulty or perception errors, the shaping signal remains silent, effectively reverting the optimization to a sparse-reward regime. Thus, while MiRA excels at extending the ``competence boundary'' of agents that can start a task, it implies that future work must still address the cold start exploration problem in environments where even the first milestone is hard to reach.

\paragraph{Towards Self-Evolving Autonomous Agents.}
Perhaps the most compelling implication of our results is the feasibility of a closed-loop, self-improving training cycle. In our experiments, a small ensemble of capable models collectively played complementary roles: a teacher model (Gemini-2.5-pro) generated and verified subgoals, a lightweight model (Gemini-2.5-flash) handled curriculum filtering and similarity matching, an ORM judged terminal success, and the trained agent (Gemma3-12B) executed and improved across phases. Looking forward, consolidating these roles into a single self-improving model is a natural next step toward truly \textbf{self-evolving agents} capable of recursive improvement. By integrating the planning, diagnosing, and training processes into a singular, cohesive loop, future systems could autonomously continually expand their capabilities without human intervention, transforming the static ``train-then-deploy'' paradigm into a lifelong learning process.

\paragraph{Limitations and broader impacts.}
Our experiments cover both the curated WebArena-Lite tasks and the broader WebArena suite, with MiRA applied to Gemma3-12B and Llama3.1-8B. These results do not establish transfer to independent websites, task distributions, or model families beyond those tested. The limited exact-equivalence F1 shows that milestones can miss successful alternative paths or introduce noisy supervision, and high shaping weights can harm final success. The subgoal generation pipeline depends on a capable teacher model, which concentrates methodological trust in proprietary systems and may introduce distributional bias from that teacher's training data. Web-navigation agents in general carry dual-use potential: while they can reduce repetitive labor and improve accessibility, the same capabilities could enable scraping, spam generation, or circumvention of access controls. We run experiments in sandboxed WebArena environments without interaction with live websites.

\paragraph{Future Directions.}
While our framework establishes a strong baseline, several dimensions warrant further exploration to generalize this approach. First, regarding milestone synthesis: we aim to transition from heuristic prompts to \emph{learnable or hierarchical subgoal generators}. Such models could dynamically tailor the granularity of milestones for knowledge-sparse domains where static linear ramps are insufficient. Second, we see promise in refining the shaping mechanism itself; future work should explore \emph{non-linear progress estimation} that accounts for the varying difficulty of distinct subgoals rather than treating them uniformly. Finally, we propose investigating \emph{signal annealing strategies}. In this regime, subgoals would serve as temporary ``warm-up'' scaffolding that is gradually withdrawn as training progresses, ensuring that the final policy relies on the true task objective rather than over-optimizing for auxiliary rewards.

\newpage
\section*{NeurIPS Paper Checklist}
\begin{enumerate}

\item {\bf Claims}
    \item[] Question: Do the main claims made in the abstract and introduction accurately reflect the paper's contributions and scope?
    \item[] Answer: \answerYes{}
    \item[] Justification: The abstract and Section~\ref{sec:intro} explicitly state the two contributions: (1) a validated subgoal generation procedure with AUROC 0.84 and Kendall's $\tau = 0.4585$ (Section~\ref{sec:sg_gen}), and (2) the MiRA RL fine-tuning framework (Section~\ref{sec:mira_rl}). The reported success rate gains (Gemma3-12B: $6.4\% \to 43.0\%$) match the experimental results in Table~\ref{tab:main_results}, and all comparisons are scoped to the WebArena-Lite benchmark.

\item {\bf Limitations}
    \item[] Question: Does the paper discuss the limitations of the work performed by the authors?
    \item[] Answer: \answerYes{}
    \item[] Justification: Limitations are discussed in the Extended Discussion (Appendix~\ref{sec:apdx_discussion}), including the ``Solid Grounding'' trade-off where shaping signals require reachable initial subgoals (leading to cold-start issues in high-difficulty settings), the reliance on a capable teacher model for subgoal generation, and the current use of linear progress interpolation as opposed to non-linear difficulty-aware estimation. 

\item {\bf Theory assumptions and proofs}
    \item[] Question: For each theoretical result, does the paper provide the full set of assumptions and a complete (and correct) proof?
    \item[] Answer: \answerYes{}
    \item[] Justification: The paper's main theoretical claim---the equivalence between Potential-Based Reward Shaping and $\Phi$-initialization in the goal-conditioned setting (Section~\ref{sec:preliminary})---is accompanied by a full induction proof in Appendix~\ref{sec:apdx_proof}, adapted from \citet{wiewiora2003potential}. The derivation of the optimal policy form $\pi^*$ used in the MSE objective is given in Appendix~\ref{sec:apdx_RL_adv}, and the MSE-vs-KL comparison derivation is in Appendix~\ref{sec:apdx_mse_vs_kl}. All assumptions (goal-conditioned MDP, state-only potential function, $\arg\max$-invariance of constant shifts) are stated explicitly.

\item {\bf Experimental result reproducibility}
    \item[] Answer: \answerYes{}
    \item[] Justification: The training pipeline (curriculum outer loop in Algorithm~\ref{alg:mira_curriculum}, inner MiRA-RL loop in Algorithm~\ref{alg:mira-rl}, potential critic training in Algorithm~\ref{alg:potential_critic_final}), the base model (Gemma3-12B), the benchmark (WebArena-Lite, 165 tasks), the system prompts (Appendix~\ref{sec:apdx_exp}, Figures~\ref{fig:system-prompt}, \ref{fig:orm-prompt}, \ref{fig:proprietary-prompt}, \ref{fig:subgoal-prompt}), and all hyperparameters (Table~\ref{tab:hyperparams}) are fully documented. The subgoal generation procedure uses publicly described prompting with Gemini-2.5-pro. All baselines (DigiRL, WebRL, AWR) are reproduced on the same SFT checkpoint for fair comparison.

\item {\bf Open access to data and code}
    \item[] Answer: \answerNo{}
    \item[] Justification: The training code and model checkpoints are not released at submission time due to ongoing internal review. The benchmark is publicly available at \url{https://github.com/THUDM/VisualAgentBench, https://github.com/web-arena-x/webarena}, and the base models (Gemma3-12B, Llama3.1-8B) are publicly released by their respective providers. We provide full algorithmic descriptions, hyperparameters, and prompts in the appendix to enable independent reimplementation.

\item {\bf Experimental setting/details}
    \item[] Answer: \answerYes{}
    \item[] Justification: Full experimental details are provided: the training protocol (Section~\ref{sec:setup}) describes initialization from shared SFT checkpoints, the 6-round curriculum for DigiRL/WebRL/MiRA, and the evaluation setup; Appendix~\ref{sec:apdx_exp} contains the complete hyperparameter table (Table~\ref{tab:hyperparams}) covering learning rates, schedulers, batch sizes, discount factors, and training epochs for all methods (SFT, AWR, DigiRL, WebRL, MiRA); observation and action spaces are specified in Appendix~\ref{sec:apdx_agent_design}; filtering thresholds (perplexity range $[1/0.9, 1/0.5]$, $\alpha=0.3$ for shaping) are reported alongside their ablations in Appendix~\ref{sec:apdx_ablation}.

\item {\bf Experiments compute resources}
    \item[] Answer: \answerYes{}
    \item[] Justification: Compute infrastructure details are provided in Appendix~\ref{sec:apdx_exp}: all experiments were conducted on a single node of 8 high-memory NVIDIA A100 GPUs (80GB each). A complete multi-phase MiRA training run on Gemma3-12B takes approximately three days of wall-clock time, and full WebArena-Lite evaluation takes ${\sim}40$ minutes per run. Ablation and multi-seed experiments were scheduled on the same hardware.

\item {\bf Code of ethics}
    \item[] Answer: \answerYes{}
    \item[] Justification: The research conforms with the NeurIPS Code of Ethics. The work uses publicly released models (Gemma3-12B, Llama3.1-8B) and a public benchmark (WebArena-Lite) built on self-hosted sandboxed web environments. No human subjects, no crowdsourced annotations beyond the 40-example validation performed by the authors, and no deployed system interacting with real users or external websites.

\item {\bf Broader impacts}
    \item[] Answer: \answerYes{}
    \item[] Justification: Potential positive impacts include improved accessibility tooling for users with disabilities and reduced manual overhead in repetitive web tasks. Potential negative impacts include misuse of automated web agents for scraping, spam generation, or circumvention of rate-limiting mechanisms. We discuss these considerations alongside limitations in Appendix~\ref{sec:apdx_discussion}. 

\item {\bf Safeguards}
    \item[] Answer: \answerNA{}
    \item[] Justification: The paper does not release new high-risk pre-trained models, image generators, or scraped datasets. The base models used (Gemma3-12B, Llama3.1-8B) were publicly released by their respective providers with their own safeguards. No new dataset beyond synthetic subgoal annotations (generated over the public WebArena-Lite benchmark) is released.

\item {\bf Licenses for existing assets}
    \item[] Answer: \answerYes{}
    \item[] Justification: WebArena~\citep{zhou2023webarena} and WebArena-Lite~\citep{liu2024visualagentbench} are cited and used under their respective open-source licenses. Base models Gemma3-12B and Llama3.1-8B are used under their published license terms. Baselines DigiRL~\citep{bai2024digirl}, WebRL~\citep{qi2024webrl}, and AWR~\citep{peng2019advantage} are cited and reproduced from their original implementations. Gemini-2.5-pro and other proprietary models are accessed through their official APIs in accordance with published terms of service.

\item {\bf New assets}
    \item[] Answer: \answerNA{}
    \item[] Justification: The paper does not release new assets (models, datasets, or benchmarks) at submission. Subgoal annotations are generated from the existing benchmark via prompting; the full prompts used are documented in Appendix~\ref{sec:apdx_exp} (Figure~\ref{fig:subgoal-prompt}) to enable regeneration.

\item {\bf Crowdsourcing and research with human subjects}
    \item[] Answer: \answerNA{}
    \item[] Justification: The paper does not involve crowdsourcing or research with human subjects. The validation of the failure classifier (Appendix~\ref{sec:apdx_failure}, Table~\ref{tab:validation_scores}) used examples labeled internally by the authors, not by external crowd workers.

\item {\bf Institutional review board (IRB) approvals or equivalent for research with human subjects}
    \item[] Answer: \answerNA{}
    \item[] Justification: The paper does not involve research with human subjects, so no IRB approval is required.

\item {\bf Declaration of LLM usage}
    \item[] Answer: \answerYes{}
    \item[] Justification: LLMs are a central component of the research methodology rather than a writing aid. Gemini-2.5-pro is used as the teacher model for subgoal generation (Section~\ref{sec:sg_gen}, Figure~\ref{fig:subgoal-prompt}) and as the AutoRater/SubGoal Checker during trace annotation. Gemini-2.5-flash is used for the automated failure analyzer (Appendix~\ref{sec:apdx_failure}) and for semantic similarity resampling in the curriculum generation loop (Appendix~\ref{apdx:iterative_rl}). An Outcome Reward Model trained via WebRL~\citep{qi2024webrl} serves as the final task-level judge. Full prompts for all LLM roles are provided in Appendix~\ref{sec:apdx_exp}.

\end{enumerate}

\end{document}